%% file: main.tex
\input{preamble/preamble.tex}

\input{preamble/commands.tex}		
\input{preamble/math_commands.tex}	
\input{preamble/colors.tex}			
\usepackage[accepted]{preamble/icml2025} 

\makeatletter
\renewcommand{\printAffiliationsAndNotice}[1]{%
\stepcounter{@affiliationcounter}%
{\let\thefootnote\relax\footnotetext{%
\hspace*{-\footnotesep}\ifdefined\isaccepted #1\fi%
\forloop{@affilnum}{1}{\value{@affilnum} < \value{@affiliationcounter}}{%
\textsuperscript{\arabic{@affilnum}}%
\ifcsname @affilname\the@affilnum\endcsname
\csname @affilname\the@affilnum\endcsname
\else
{\bf AUTHORERR: Missing \textbackslash{}icmlaffiliation.}%
\fi
}.\enspace%
\ifdefined\icmlcorrespondingauthor@text
Correspondence to: \icmlcorrespondingauthor@text.%
\else
{\bf AUTHORERR: Missing \textbackslash{}icmlcorrespondingauthor.}%
\fi
\par\smallskip %
\textit{Preprint.}%
}}}
\makeatother

\icmltitlerunning{Low-Rank Filtering \& Smoothing for Sequential Deep Learning}

\begin{document}

\twocolumn[
  \icmltitle{Low-Rank Filtering and Smoothing for Sequential Deep Learning}

  \icmlsetsymbol{equal}{*}
  
  \begin{icmlauthorlist}
    \icmlauthor{Joanna Sliwa}{tue}
    \icmlauthor{Frank Schneider}{tue}
    \icmlauthor{Nathanael Bosch}{tue}
    \icmlauthor{Agustinus Kristiadi}{vec}
    \icmlauthor{Philipp Hennig}{tue}
  \end{icmlauthorlist}

  \icmlaffiliation{tue}{\aicenter, \UoT, Tübingen, Germany}
  \icmlaffiliation{vec}{\vectorinst, Toronto, Canada}

  \icmlcorrespondingauthor{Joanna Sliwa}{joanna.sliwa@uni-tuebingen.de}

  \icmlkeywords{Machine Learning, ICML, Sequential Learning, Continual Learning, Filtering, Smoothing, Bayesian Inference, Neural Networks}

  \vskip 0.3in
]

\printAffiliationsAndNotice{}  

\begin{abstract}
	Learning multiple tasks sequentially requires neural networks to balance retaining knowledge, yet being flexible enough to adapt to new tasks.
	Regularizing network parameters is a common approach, but it rarely incorporates prior knowledge about task relationships, and limits information flow to future tasks only.
	We propose a Bayesian framework that treats the network's parameters as the state space of a nonlinear Gaussian model, unlocking two key capabilities:
	(1) A principled way to encode domain knowledge about task relationships, allowing, e.g., control over which layers should adapt between tasks.
	(2) A novel application of Bayesian smoothing, allowing task-specific models to also incorporate knowledge from models learned later.
	This does not require direct access to their data, which is crucial, e.g., for privacy-critical applications.
	These capabilities rely on efficient filtering and smoothing operations, for which we propose diagonal plus low-rank approximations of the precision matrix in the Laplace approximation (LR-LGF).
	Empirical results demonstrate the efficiency of LR-LGF and the benefits of the unlocked capabilities.
\end{abstract}


\section{Introduction}
\label{sec:intro}
The central challenge in sequential deep learning is preserving previously acquired knowledge without compromising flexibility.
In \textit{continual learning}, models encounter new tasks one by one, that they need to learn while preserving knowledge from all previous tasks.
Despite the model's capacity to learn all tasks \emph{simultaneously}, sequential training often leads to partial ``forgetting'' of past tasks.
Divergence between a sequentially \& simultaneously learned model is referred to as \textit{catastrophic forgetting} \citep{McCloskey1989}.
Common mitigation strategies constrain the network to encourage changes that maintain prior task performance, \eg~via weight-space regularization.
However, an excessive focus on retention can lead to an inability to adapt to new tasks, termed \textit{loss of plasticity} \citep{Dohare2023}.
Achieving a good trade-off requires a quantification of uncertainty---knowing what remains relevant, and which model parts can be adapted.

Building on recent work (e.g. \citealp{Ritter2018}; \citealp{chang2022on}; \citealp{chang2023lowrank}), we treat the weight space of the deep network as the state space of a nonlinear Gaussian model.
The main advantage of this approach is that it maps the complex problem of sequential learning to the well-understood formalism of Bayesian inference.
This enables both efficient computational methods, \eg Bayesian filtering and smoothing, as well as clear conceptual understanding of the state space model's components, \eg by providing principled ways to include domain knowledge about the sequence of tasks.

\begin{figure*}
	\centering
	\resizebox{0.95\textwidth}{!}{\input{figs/visual_abstract.tex}}
	\caption{
		\textbf{An efficient weight-space Laplace--Gaussian filter (\colorarrow{TUlightblue}\,, \colorarrow{TUpurple}) and smoother (\colorreversearrow{TUlightred}) for sequential deep learning.}
		We treat the neural network's parameters as a nonlinear Gaussian state-space model and perform efficient inference using diagonal plus low-rank Laplace--Gaussian filtering and smoothing.
		During the update step (\colorarrow{TUpurple}) we train the neural network on the current task using the parameter covariance as a regularizer, then approximate the posterior distribution with a diagonal plus low-rank Laplace approximation.
		The predict step (\colorarrow{TUlightblue}) adds noise to the model parameters, where the noise covariance $\mQ$ can be used to model the type of shift between tasks.
		Smoothing (\colorreversearrow{TUlightred}) allows for training task-specific model parameters \(\params_t\) that are informed by \emph{all} tasks, without additional training.
	}
	\label{fig:visual_abstract}
\end{figure*}

\textbf{Contributions:} We propose a grounded framework for sequential neural network training on related tasks, as in continual, transfer, or online learning, based on a Laplace--Gaussian filter and smoother allowing the following capabilities:
\begin{enumerate}[left=\parindent]
	\item \textbf{Changes between tasks can be described with established concepts from Bayesian inference.}
	A Bayesian filter can incorporate domain knowledge about task relationships. Specifically, we study how structured uncertainties can be integrated via the process noise matrix. This matrix models (possible) stochastic drifts between tasks. We demonstrate that, one can integrate prior domain knowledge about how model parameters are likely to change in the \textit{subsequent} related tasks (\Cref{sec:exp_Q}). For example, it is straightforward to include information indicating that primarily the model's upper (or lower) layers should change between tasks.
	\item \textbf{Previous tasks can be efficiently informed by knowledge of subsequent tasks.}
	We examine the benefits of Bayesian smoothing, and sequentially train \emph{task-specific models} that are informed by \emph{all} tasks, without requiring renewed access to the data. We demonstrate that smoothing can significantly boost the performance in initial tasks (\Cref{sec:exp_smoother}). This might be critical in low-data or privacy-critical applications and extends our framework to fields outside of continual learning.
	\item \textbf{All required operations can be computed efficiently via diagonal plus-low rank approximations.}
	As part of the filter (\Cref{sec:kalman}), we employ a weight-space regularizer.
	Naive filtering or smoothing implementations---even with approximations---would incur prohibitive computational costs.
	We resolve this by utilizing the Laplace approximation \citep[\eg][]{MacKay1992,daxberger2022laplace} with the generalized Gauss-Newton (GGN) matrix \citep{Schraudolph2002}.
	By exploiting the low-rank structure of the GGN, we construct \textit{diagonal plus low-rank approximations} of the precision matrices via truncated SVDs.
	Leveraging this structure, all required filtering and smoothing operations can be rewritten and computed efficiently (\Cref{sec:method:lrlgf}).
	We provide an open-source \textsc{JAX} implementation of this efficient diagonal plus low-rank Laplace--Gaussian filter, termed LR-LGF.\footnote{See \url{https://github.com/a3724/lr-lgf}}
	We demonstrate that this diagonal plus-low rank approximation provides a reasonable yet cheap approximation, while enabling efficient filtering \& smoothing (\Cref{sec:exp_GGN}).
\end{enumerate}


\section{Background}
\label{sec:background}

\textbf{Notation:}
We consider supervised learning with a \dataset $\sD\!=\!\left\{(\vx_i, \vy_i) | i\!=\!1, \dots, \numtraindata
\right\}$ containing training inputs $\vx_i$ and outputs $\vy_i$.
The objective is to find parameters $\params\!\in\!\R^{\numparams}$ of a deep network $f_{\params}$ that minimize a given loss, \ie $\params^{\ast}\!=\!\argmin_{\params} \Loss(\vy;f_{\params}(\vx))\!=\!\argmin_\params \Loss(\params,\sD)$.
The first and second derivatives of the loss with respect to the parameters are represented by the gradient $\vg(\params)\!=\!\nabla_\params \Loss(\vy;f_{\params}(\vx))\!\in\!\R^{\numparams}$ and the Hessian $\mH(\params)\!=\!\nabla_\params^2 \Loss(\vy;f_{\params}(\vx))\!\in\!\R^{\numparams\!\times\!\numparams}$.

\textbf{Continual learning:}
While no unified definition exists (\Cref{app:CL_definition}), we describe continual learning as sequentially processing a series of tasks while maintaining good performance on \emph{all} tasks concurrently.
Specifically, for a sequence of tasks $t\!\in\!\{t_1,\dots,t_T\}$, described by \datasets $\sD_t$, we prioritize the \textit{average} performance of a single model $f_{\params}$ across all tasks so far, measured by the average loss (or accuracy) $\frac{1}{T} \sum_{t} \Loss(\params, \sD_t)$.
Crucially, tasks are experienced sequentially, without access to data from previous tasks.
To achieve the required balance between model flexibility \& rigidity, we adopt a regularization-based approach (see \Cref{sec:related_work}), wherein we train the model on task $t$ using a non-isotropic $\ell_2$ regularizer.
This regularizer represents prior knowledge and thus constraints from previous tasks \citep[see][]{Kirkpatrick2017, Ritter2018}.
It effectively encodes which weights should remain unchanged to maintain performance on previous tasks and which can be adjusted to perform well on the new task.
To implement this, we utilize the Laplace approximation of the Bayesian posterior over the model's parameters, such that the posterior of the previous task becomes the prior for the current one.

\textbf{Laplace approximations:}
The Bayesian posterior over the model's parameters, $p(\params \mid \sD)$, describes the current belief over the specific values of each parameter.
It thus reflects (un)certainty about each parameter's value and identifies which parameters still offer flexibility to learn new tasks.
The \textit{Laplace approximation} \citep[\eg][]{MacKay1992,daxberger2022laplace} constructs a local Gaussian approximation to this typically intractable posterior.
It arises from a second-order Taylor expansion of the loss around the maximum a posteriori (MAP) estimate, \ie the trained $\params^{\ast}$, as $\Loss(\params, \sD) \!\approx\! \Loss(\params^{\ast}, \sD) + \tfrac{1}{2} \left(\params - \params^{\ast}\right)^\top \mH(\params^{\ast})  \left(\params - \params^{\ast}\right)$, yielding a Gaussian distribution $p(\params\mid\sD) \!\approx\! \mathcal{N}(\params;\params^{\ast}, \mH^{-1}(\params^{\ast}))$ known as the Laplace approximation.
When used as a weight-space regularizer with strength $\lambda$, this results in the regularized loss for task $t$ as $\RegLoss(\params, \sD_t)\!=\!\Loss(\params, \sD_t)+ \tfrac{\lambda}{2}\! \left(\params - \params_{t-1}^{\ast}\right)^\top \!\mH_{t-1}(\params_{t-1}^{\ast}) \! \left(\params - \params_{t-1}^{\ast}\right)$.
Intuitively, we prefer solutions for task $t$ to remain near the previously trained (MAP) parameters $\params_{t-1}^{\ast}$.
High-curvature parameters (where changes degrade prior performance) are preserved, while allowing adaptation in low-curvature directions.

\textbf{Generalized Gauss-Newton:}
Since the Hessian is the second derivative of a composition of two functions, $\Loss$ and $f$, we can rewrite it with $\mJ\!\in\!\R^{\numparams\times\numclasses}$ and $\hat{\mH}\!\in\!\R^{\numclasses\times\numclasses}$ as
\begin{equation}
	\!\!\!\mH(\params)
	= \frac{\partial f}{\partial \params} \frac{\partial^2 \Loss}{\partial f^2} \frac{\partial f}{\partial \params}^{\top}\!+ \frac{\partial^2 f}{\partial \params^2} \frac{\partial \Loss}{\partial f}
	\defeq \mJ \hat{\mH} \mJ^\top\!+ \frac{\partial^2 f}{\partial \params^2} \frac{\partial \Loss}{\partial f}\,
\end{equation}
where \(\numclasses\) denotes the neural net's output dimension.
The generalized Gauss-Newton (GGN) matrix is defined as the first term of this expression, \(\mJ \hat{\mH} \mJ^\top\)
\citep{Schraudolph2002}.
Since typically \(\numclasses\!\ll\!\numparams\),
the GGN is low-rank, can be stored in \(\bigO( \numparams \numclasses)\),
and is guaranteed to be positive semi-definite.
In addition, the GGN can be computed using mini-batches, \ie $\mJ \hat{\mH} \mJ^\top\!=\!\sum_{b=1}^B \mJ_t^{(b)} \hat{\mH}_t^{(b)} ( \mJ_t^{(b)} )^\top\!$, where \(\mJ_t^{(b)}\!\in\!\R^{\numparams \times \numclasses}\) is the network's Jacobian with respect to its parameters and \(\hat{\mH}_t^{(b)}\!\in\!\R^{\numclasses \times \numclasses}\) is the Hessian of the loss with respect to the neural network outputs,
for the \(b\)-th mini-batch.


\section{A Bayesian Inference Framework for Sequential Learning}
\label{sec:kalman}

We consider a sequence of tasks
$t\!=\!1, 2, \ldots, T$
with corresponding datasets
$\sD_t$.
The goal is to compute a posterior distribution over the net's parameters, given the data of all prior tasks,
\ie $p(\params_t \mid \sD_{1:t})$.
We are also interested in the posterior distribution for task $t$ given \emph{all} available \datasets, \ie $p(\params_t \mid \sD_{1:T})$.
We formulate these distributions as the \emph{filtering} \& \emph{smoothing} distributions in a suitable Gaussian state-space model and develop an approximate inference algorithm to efficiently estimate these distributions.

Sequential training can be phrased as a \emph{Bayesian state estimation} problem, where the parameters of the network are treated as the state of a state-space model of the form
\begin{subequations}
	\label{eq:lgf:state_space}
	\begin{align}
		\label{eq:lgf:state_space:transition}
		\!\!\!\!\!\text{Transition model:} &  &\!\!\!\!\! &p(\params_{t+1} \mid \params_t) = \mathcal{N}(\params_{t+1}; \params_t, \mQ),                             \\
		\label{eq:lgf:state_space:likelihood}
		\text{Likelihood:}       &  &\!\!\!\!\! &p(\sD_t \mid \params_t)\propto \exp\!\left(\!-\frac{1}{\lambda} \Loss(\params_t, \sD_t)\! \right)\!.
	\end{align}
\end{subequations}
The un-normalized likelihood
\(p(\sD_t \mid \params_t)\)
encodes the supervised learning task on the dataset \(\sD_t\), defined by the loss function \(\Loss\),
scaled by a factor \(1/\lambda \!\in\! \R_+\) which controls the regularization strength.
The prior transition density
\(p(\params_{t+1} \mid \params_t)\)
describes a prior belief over the parameter change from task \(t\) to \(t\!+\!1\), with diagonal Gaussian noise covariance \(\mQ\!\in\!\R^{\numparams\!\times\!\numparams}\).
Then, computing the posterior over the weights given the data up to task \(t\), that is \(p(\params_t \mid \sD_{1:t})\), is known as Bayesian filtering \citep{Sarkka_Svensson_2023}.

In state-space models, the posterior distribution \(p(\params_t \mid \sD_{1:t})\) can be computed recursively using the so-called general Bayesian filtering equations \citep{Sarkka_Svensson_2023}:
\begin{align}
	\label{eq:gbf:predict}
	&\text{Predict step:}\notag \\
	&p(\params_{t} \mid \sD_{1:t-1}) = \int p(\params_{t} \mid \params_{t-1}) p(\params_{t-1} \mid \sD_{1:t-1}) \dd \params_{t-1}, \\
	\label{eq:gbf:update}
	&\text{Update step:}\;
	p(\params_{t} \mid \sD_{1:t}) \propto\  p(\sD_{t} \mid \params_{t}) p(\params_{t} \mid \sD_{1:t-1}).
\end{align}
These equations demonstrate the value of Bayesian filtering \& smoothing for sequential learning as they describe an exact, recursive procedure to learn from a sequence of \datasets.
However, the exact Bayesian predict \& update steps are intractable for all but the simplest state-space models.
We will show how to efficiently approximate them with a diagonal plus low-rank Laplace--Gaussian filtering algorithm.

\subsection{Laplace--Gaussian Filtering}
\label{sec:method:lgf}

The Laplace--Gaussian filter (LGF)
\citep{koyama2010}
approximates the posterior with Gaussian distributions
\(
  {p(\params_t \mid \sD_{1:t}) \!\approx\! \mathcal{N}(\params_t; \vm_t, \mC_t)}.
\)
This approach is commonly known as \emph{Gaussian filtering},
and includes many well-known algorithms such as the extended, and the unscented Kalman filter
\citep{jazwinski2007stochastic,Sarkka_Svensson_2023,Uhlmann2000}.
The predict and update steps of the LGF  are as follows.

\textbf{Predict step:}
Since we assume
\(p(\params_{t-1} \mid \sD_{1:t-1})\)
to be Gaussian with mean \( \vm_{t-1} \) and covariance \( \mC_{t-1} \),
and since
\(p(\params_{t} \mid \params_{t-1})\)
is Gaussian as given in \cref{eq:lgf:state_space:transition},
the exact predictive distribution as in \cref{eq:gbf:predict} is also Gaussian, with
mean \(\vm_{t}^- \!=\! \vm_{t-1}\) and covariance \(\mC_{t}^- \!=\! \mC_{t-1} + \mQ\).
This is exactly equivalent to the predict step of a standard Kalman filter
\citep{kalman1960,KalmanBucy1961}.

\textbf{Update step:}
The exact update of \cref{eq:gbf:update} is intractable as the likelihood model is not only non-linear, but also non-Gaussian and un-normalized.
Therefore, we Laplace-approximate the filtering distribution with a Gaussian distribution
\(
p(\params_{t} \mid \sD_{1:t}) \!\approx\! \mathcal{N}(\params_{t}; \vm_{t}, \mC_{t}),
\)
where the mean $\vm_{t}$ is the mode of the filtering distribution, and $\mC_{t}$ is the inverse Hessian.
This is known as Laplace--Gaussian filtering \citep{koyama2010}.
More precisely, we define the regularized loss \(\RegLoss_t(\params_{t})\) by taking the negative log of the un-normalized posterior and discarding constant terms, as
\begin{align}
  \RegLoss_t(\params_{t})
  & \!\defeq\!
  \Loss(\params_{t}, \sD_{t}) + \tfrac{1}{2}\! \left( \params_{t} - \vm_t^- \right)^{\!\top} \! \left( \mC_t^- \right)^{\!-1}\! \left( \params_{t} - \vm_t^- \right) \notag \\
  & \;\propto\!
  - \log p(\params_{t} \mid \sD_{1:t}).
\end{align}
Then, the mean and covariance of the Laplace-approximated filtering distribution are given by $\vm_{t} \!=\! \argmin_{\params} \RegLoss_t(\params)$ and $\mC_{t} \!=\! \left( \nabla^2 \RegLoss_t(\vm_{t}) \right)^{-1}$.
The first term is computed via optimization and the second term can be further decomposed
into the loss Hessian and the prior covariance $\mC_t^-$, as
\begin{align}
  \label{eq:lgf:update:cov}
  \mC_{t} = \left( \lambda \nabla^2 \Loss(\params, \sD_{t})\big|_{\params=\vm_{t}} + \left( \mC_t^{-} \right)^{-1} \right)^{-1},
\end{align}
which follows from the linearity of the Hessian operator. 
Following \citet{Ritter2018}, and do not include $\lambda$ in the loss function but the covariance to prevent the regularization strength from propagating through recursion.
In summary, the update step essentially consists of training the network on the new task using a regularized loss function to avoid forgetting previous tasks, and then updating the covariance of the filtering distribution based on the curvature of the un-regularized loss function and the prior covariance.

However, this remains impractical for deep learning due to the dense covariance and Hessian matrices being of prohibitive size.
We resolve this with low-rank approximations.

\subsection{Efficient Low-Rank Laplace--Gaussian Filtering with the GGN}
\label{sec:method:lrlgf}

The main bottleneck of the previous algorithm in the context of deep learning lies in computing and storing
the dense covariance matrices \(\mC_t, \mC_t^- \!\in\! \R^{\numparams\!\times\!\numparams}\)
and the exact Hessian.
We resolve both issues together by formulating the algorithm to track only diagonal plus low-rank matrices and by using the GGN as a low-rank Hessian approximation, \ie \( \mH_t \!\approx\! \sum_{b=1}^B \mJ_t^{(b)} \hat{\mH}_t^{(b)} ( \mJ_t^{(b)} )^\top \).
More precisely, we track diagonal plus low-rank approximations of the \emph{precision} matrices, \ie the inverse covariance matrices \(\mP_t \!=\! \mC_t^{-1}\), of the form
\(
  \mP_t = \mD_t + \mU_t \mSigma_t \mU_t^\top,
\)
where \(D_t \!\in\! \R^{\numparams \!\times\! \numparams}\) is diagonal,
\(U_t \!\in\! \R^{\numparams \!\times\! \numrank}\) is a tall,
and \(\Sigma_t \!\in\! \R^{\numrank \!\times\! \numrank}\) a dense matrix,
with rank \(\numrank \!\ll\! \numparams\).
This resolves the storage and computational issues related to the covariance matrices' size:
Storing the diagonal and low-rank matrices requires only \(\numparams + \numparams \numrank + \numrank^2 \!\ll\! \numparams^2\) parameters,
and the cost of matrix-vector products with the precision matrix is reduced from \(\mathcal{O}(\numparams^2)\) to \(\mathcal{O}(\numparams \numrank + \numrank^2)\).

What remains is to demonstrate how to preserve the diagonal plus low-rank structure in the predict and update steps.

\textbf{Diagonal plus low-rank predictive precision:}
Given a diagonal plus low-rank precision matrix
\(\mP_{t-1} \!=\! \mD_{t-1} + \mU_{t-1} \mSigma_{t-1} \mU_{t-1}^\top\)
and a diagonal process noise covariance \(\mQ\),
the predicted precision matrix is also diagonal plus low-rank,
with parameters \(\left( \mD_t^-, \mU_t^-, \mSigma_t^- \right)\) given by
\begin{subequations}
	\label{eq:lgf:low_rank_predict}
	\begin{align}
	\label{eq:lgf:low_rank_predict:D}
    \mD_t^-     &\!=\! \left( \mQ + \mD_{t-1}^{\!-1} \right)^{-1}\!\!, \; \\
	\label{eq:lgf:low_rank_predict:U}
    \mU_t^-     &\!=\! \left( \mQ + \mD_{t-1}^{-1} \right)^{\!-1} \mD_{t-1}^{-1} \mU_{t-1}, \\
	\label{eq:lgf:low_rank_predict:Sig}
    \mSigma_t^- & \!=\! \Big(
    \mSigma_{t-1}^{-1} + \mU_{t-1}^\top \mD_{t-1}^{-1} \mU_{t-1}  \\
    &\! - \mU_{t-1}^\top \mD_{t-1}^{-\top} \left( \mQ_{t-1}+ \mD_{t-1}^{-1} \right)^{-1} \mD_{t-1}^{-1} \mU_{t-1} \notag
    \Big)^{\!-1}.
    \end{align}
\end{subequations}
This follows from applying the Woodbury matrix identity twice;
full derivation in \cref{app:low_rank_predict}.

\textbf{Diagonal plus low-rank filtering precision:}
The filtering precision matrix \(\mP_t\) is a sum of the predicted precision matrix and the Hessian of the loss function for task \(t\) (\cref{eq:lgf:update:cov}).
By approximating the full Hessian with the GGN matrix, we can write the filtering precision matrix as
\begin{equation}
  \mP_t = \mD_t^- + \mU_t^- \mSigma_t^- \mU_t^{-\top} + \sum_{b=1}^B \mJ_t^{(b)} \hat{\mH}_t^{(b)} \left( \mJ_t^{(b)} \right)^\top.
\end{equation}
To see that this is a diagonal plus low-rank matrix, with increased rank, we denote matrix square-roots by \(\mA^{\!\sfrac{1}{2}}\), with
\(\mA^{\!\sfrac{1}{2}} (\mA^{\!\sfrac{1}{2}})^{\!\top} \!=\! \mA\),
and define \(\mW_t \!\in\! \R^{\numparams \!\times\! (\numrank + B \numclasses)}\) as
\begin{equation}
	\label{eq:W_t}
  \mW_t \!\defeq\! \begin{bmatrix}
    \mU_t^- \!\left( \mSigma_t^-\right)^{\!\sfrac{1}{2}} &\!\! \mJ_t^{(1)}\!\left( \hat{\mH}_t^{(1)} \right)^{\!\sfrac{1}{2}} &\!\! \cdots &\!\! \mJ_t^{(B)}\!\left( \hat{\mH}_t^{(B)} \right)^{\!\sfrac{1}{2}}
  \end{bmatrix}.
\end{equation}
The filtering precision matrix can then be written \(
  \mP_t = \mD_t^- + \mW_t \mW_t^\top,
\)
but with increased rank \(\numrank \!+\! B \numclasses\).
To prevent rank inflation, we compress the matrix
\(\mW_t\) by performing a truncated singular value decomposition (SVD)
of rank \(\numrank\)
to obtain
\(\mW_t \!\approx\! \tilde{\mU}_t \tilde{\mSigma}_t \tilde{\mV}_t^\top\),
with \(\tilde{\mU}_t \!\in\! \R^{\numparams \!\times\! \numrank}\),
\(\tilde{\mV}_t \!\in\! \R^{\numrank \!\times\! \numparams}\),
and diagonal \(\tilde{\mSigma}_t \!\in\! \R^{\numrank \!\times\! \numrank}\).
Then, the filtering precision matrix is
\begin{align}
  \mP_t &= \mD_t^- + \tilde{\mU}_t \tilde{\mSigma}_t \tilde{\mV}_t^\top \tilde{\mV}_t \tilde{\mSigma}_t \tilde{\mU}_t^\top \notag \\
  &= \mD_t^- + \tilde{\mU}_t \tilde{\mSigma}_t^2 \tilde{\mU}_t^\top.
  \eqdef \mD_t + \mU_t \mSigma_t \mU_t^\top.
\end{align}
The resulting diagonal plus low-rank approximation of the filtering precision matrix enables the Laplace--Gaussian filter (\cref{sec:method:lgf}) for continual deep learning (\cref{alg:low_rank_filtering}).

\input{figs/algorithm.tex}

\subsection{Task-Specific Models via Backwards Smoothing}
\label{sec:method:lgs}
Until now, our focus has been on computing and storing a single model trained sequentially on tasks. 
However, if the tasks differ, it may be beneficial to store \emph{task-specific} models, that are informed by all \datasets---still with the restriction that the \datasets are only observed sequentially.
For state-space models, this is known as \emph{smoothing}
\citep{rauchtungstriebel1965,Sarkka_Svensson_2023}.
Since we consider Gaussian filtering distributions, and the transition model is linear and Gaussian (\cref{eq:lgf:state_space:transition}),
the smoothing distribution is also Gaussian, i.e.\ \(p(\params_t \mid \sD_{1:T}) \!\approx\! \mathcal{N}(\params_t; \vm_t^s, \mC_t^s)\),
and its mean and covariance can be computed recursively backwards in time:
Starting with \(\vm_T^s \!=\! \vm_T\) and \(\mC_T^s \!=\! \mC_T\),
the smoothing equations are given by
\citep{rauchtungstriebel1965}
\begin{subequations}
	\label{eq:lgf:smoothing:mean-and-cov}
	\begin{align}
  	\label{eq:lgf:smoothing:mean}
  	\vm_t^s & = \vm_t + \mG_t \left( \vm_{t+1}^s - \vm_{t+1}^- \right), \\
  	\label{eq:lgf:smoothing:cov}
  	\mC_t^s & = \mC_t + \mG_t \left( \mC_{t+1}^s - \mC_{t+1}^- \right) \mG_t^\top,
	\end{align}
\end{subequations}
where \(\mG_t \!=\! \mC_t \left( \mC_{t+1}^- \right)^{-1}\) is known as the \emph{smoothing gain}.
The smoothing equations can be formulated in terms of precision matrices, and if the filtering precision matrix is diagonal plus low-rank, then the smoothing precision matrix is also diagonal plus low-rank (full derivation in \cref{app:low_rank_smoother}).
\cref{eq:lgf:smoothing:mean-and-cov} also shows why the diagonal plus low-rank structure is crucial for efficient computation of the smoothing means:
If \(\mG_t\!\in\!\R^{\numparams\!\times\!\numparams}\) were dense, the matrix-vector product would have a prohibitive computational cost of \(\mathcal{O}(\numparams^2)\), but using \(\mG_t \!=\! \left( \mP_t \right)^{-1} \mP_{t+1}^-\) and implementing the matrix vector product as a sequential product with two diagonal plus low-rank matrices, the computational cost is reduced to \(\mathcal{O}(\numparams \numrank + \numrank^2)\) and smoothing becomes feasible.


\section{Related Work}
\label{sec:related_work}

\textbf{Continual learning:} 
\citet{review_cl} categorizes approaches to continual learning as either \mbox{\textit{regularization-},} \textit{optimization-}, \textit{representation-}, \textit{architecture-}, or \textit{replay}-based (\Cref{app:CL_approaches}).
In this paper, we use a (weight) regularization-based type of approach which tackles catastrophic forgetting in neural nets by constraining the model's weight space.
This approach aims to identify weights that are important for the previous tasks, and constrains their changes in subsequent tasks.
A scalar hyperparameter $\lambda$, used on the regularizer, allows trading off performance on previous tasks vs.~the ability to learn new tasks.
\textit{Elastic Weight Consolidation} (EWC) \citep{Kirkpatrick2017} takes inspiration from neuroscience and places a quadratic constraint on the model's parameters through a regularized loss.
In each task, the regularizer consists of the sum of penalties of previous tasks where the importance of the weights is captured by the diagonal Fisher information matrix.
In contrast, \textit{Online Structured Laplace Approximations} (OSLA) \citep{Ritter2018} uses a block-diagonal K-FAC \citep{martens2020optimizing} Hessian approximation.
Their regularizer consists of a single penalty, recursively updated with the most recent log likelihood scaled by $\lambda$.
The authors observe improved performance over EWC, attributed to a more expressive Hessian approximation also capturing parameter interactions within a layer.
For the exact definitions, see \Cref{app:CL_approaches}.
Other weight regularization-based methods try to find better ways to represent the importance measure \citep{zenke2017continual}, refine the penalty \citep{8545895, 9009502}, use a expansion-renormalization approach \citep{NIPS2017_f708f064,pmlr-v80-schwarz18a}, or target the network \citep{nguyen2018variational}.
EWC and OSLA are closest to our approach since we update the precision of our penalty and scale the regularizer similarly.
However, they use different curvature approximations and neither method uses a Bayesian filter or smoother.
For related work on Gaussian Processes for CL, see \Cref{app:CL_approaches}.

\textbf{Bayesian filtering \& smoothing:}
A well-established formalism for sequential data is Bayesian filtering \& smoothing \citep{Sarkka_Svensson_2023}.
Learning neural network weights via an extended Kalman filter (EKF) was proposed already in the \textquotesingle90s \citep{singhal1988,Feldkamp1998},
but the EKF and related methods
\citep{jazwinski2007stochastic,Sarkka_Svensson_2023,Uhlmann2000}
suffer from quadratic memory and cubic computational costs, which is prohibitive for deep learning.
Therefore, various approximations have been proposed.
Diagonal EKF variants for both training and online learning of neural networks exist, but as they ignore interactions between the weights their quality is often limited
\citep{puskorius1991,chang2022on}.
Treating only the network's last layer probabilistically makes the Kalman filter tractable,
but also comes reduces expressiveness
\citep{titsias2024kalman}.
Recently, diagonal plus low-rank approximations have been proposed as a more expressive alternative and the resulting EKF variant, called LO-FI, has been shown to be effective for online learning from streaming data
\citep{chang2023lowrank}.
Building on LO-FI, we focus on continual learning, where each task's data \(\sD_t\) is observed in its entirety and performance across past tasks matters.
Therefore, instead of the EKF we use a Laplace--Gaussian filter
\citep{koyama2010}, compute the MAP estimate in each filtering step via optimization, and compute the posterior with a diagonal plus low-rank Laplace approximation.
This differs strongly from the EKF and is more akin to the more accurate \emph{iterated} EKF (IEKF)
\citep{Bell1993,Sarkka_Svensson_2023}.
Additionally, we propose the use of a smoother to improve performance on past tasks for task-specific models.


\section{Experiments}
\label{sec:experiments}

We showcase the benefits of the filtering framework by first studying how domain knowledge can be integrated via the dynamics model (\cref{sec:exp_Q}).
Next, we examine the benefits of smoothing and find that it can boost the performance of task-specific models learned ``earlier'', without renewed access to data (\cref{sec:exp_smoother}).
Lastly, we analyze our diagonal plus low-rank approximation of the GGN, used to make the filtering \& smoothing operations efficient, comparing it to other weight-space regularizers proposed in the literature (\cref{sec:exp_GGN}).

\subsection{Integrating Domain Knowledge via Q}
\label{sec:exp_Q}
\begin{figure*}[ht]
	\centering
	\includegraphics[width=0.99\linewidth]{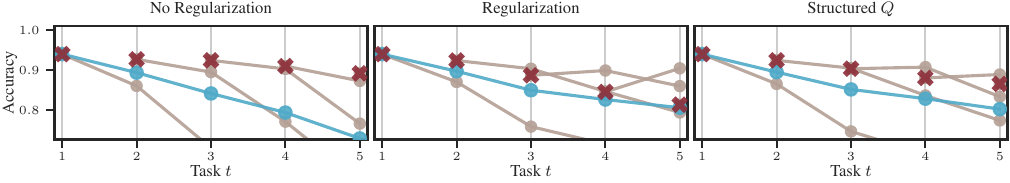}
	\caption{\textbf{The effect of $\mQ$ on the average and current task's performance.} \textit{(Left)} Without regularization, we see a significant drop in the average performance across all seen tasks (\colorline{TUlightblue}), while the performance on the current task (\colorcross{TUred}) is strong. \textit{(Center)} Adding regularization, helps boost the average performance across tasks, but to the detriment of the current task (most notably task $t\!=\!5$). However, older tasks (\colordot{TUmauve}) suffer much less from catastrophic forgetting. \textit{(Right)} Additionally using a structured $\mQ$ can boost the current task performance, while keeping the same average performance across tasks. See \cref{fig:Q_seeds} in \cref{app:exp_details_results} for a summary of the same experimental results across $8$ random seeds.}
	\label{fig:Q}
\end{figure*}

We examine the benefits of incorporating domain knowledge, \ie how tasks are related, via the dynamics model, specifically, the process noise matrix $\mQ$.
Roughly speaking, $\mQ$ adds uncertainty to the next task's model parameters (\cref{eq:lgf:state_space:transition}), describes task relationship, and thus controls the model's flexibility.
For $\mQ\!=\!0$, we assume all sub-tasks $t$ belonging to a shared meta-task, or that the optimal model parameters do not change between tasks.
In contrast, by using a structured $\mQ$, we can incorporate how we believe the tasks, and therefore the model parameters, change, \eg indicating that mostly the lower layers of the network change.
This could be uselful in Bayesian optimization, for example, in material discovery with the use of foundation models \citep{kristiadi2024soberlookllmsmaterial}.

\textbf{Setting:}
We empirically examine this on the \camelyon \dataset \citep{koh2021wilds}, which consists of images of either healthy or cancerous cells collected from different hospitals.
To showcase the benefits of $\mQ$ (and the smoother in \cref{sec:exp_smoother}), we create a continual learning task with an \emph{ordered} series of tasks that we term \textsc{Gradual \camelyon},
by gradually changing the brightness of the samples between tasks.
Task $t\!=\!1$ has the darkest and task $t\!=\!5$ has the brightest pixels.
We use a network with three convolutional and two dense layers (\cref{app:exp_details_setup} provides full details).

\textbf{Results:}
\Cref{fig:Q} illustrates that while regularization helps boost the model's average performance across tasks, it comes at the detriment of the current task's performance.
We can re-introduce flexibility by adding a non-zero process noise $\mQ$ (\cref{fig:Q} \textit{(right)}) to this regularization.
This prioritizes the current task (see also \cref{fig:Q_seeds} \textit{(right)}) while maintaining the same average accuracy (\cref{fig:Q_seeds} \textit{(center)}).
Across experiments, a structured transition noise (\ie non-zero values on the lowest layers) performs slightly better than a simpler isotropic, scalar-times-identity $\mQ$ (\cref{fig:Q_seeds}).
The structured $\mQ$ encodes an intuition that brightness changes likely affect the lower convolutional layers the most.
It demonstrates how structural domain knowledge about task relationship can be explicitly included in the filtering framework, in this case to prioritize local performance.
A non-zero $\mQ$ indicates that tasks---and thus optimal model parameters---differ.
In those cases, it may be beneficial to store \emph{task-specific models} instead of a single model for all tasks, which we will explore, via smoothing, next.

\subsection{Going Back in Time via Smoothing}
\label{sec:exp_smoother}
\begin{figure*}[t]
	\centering
	\includegraphics[width=0.99\linewidth]{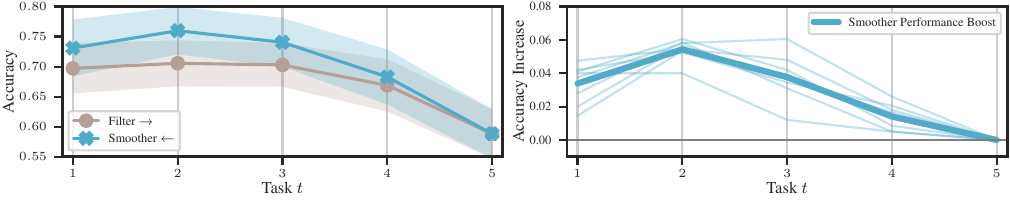}
	\caption{\textbf{Smoothing improves the performance on earlier tasks.} \textit{(Left)} Task-wise performance after filtering (\colorline{TUmauve}) or smoothing (\colorline{TUlightblue}) up to task $t$, with shaded regions representing one standard deviation across $8$ seeds. \textit{(Right)} The smoother consistently improves performance by incorporating information from all tasks without accessing their data. The thick line is the mean improvement of smoothed vs.~ the filtered model with the thin lines showing the improvements for each seed.}
	\label{fig:smoother}
\end{figure*}

We investigate the benefits of using a smoother to imbue previous tasks with the knowledge gained from later tasks.
We use the experimental setting of \cref{sec:exp_Q} with fewer data points per task.
Crucially, we now learn task-specific models $\params_t$, for each task $t\!=\!1, \dots, 5$.
As the filter computes posteriors \(p(\params_t \mid \sD_{1:t})\),
the model for the last task $\params_5$ has incorporated information from all previous tasks.
However, the model for the first task $\params_1$ is still only informed by the first task.
Using a Bayesian smoother (\cref{sec:method:lgs}), we can also update the first model with subsequent knowledge, \emph{without any additional access to any data}.
We simply update earlier model parameters based on the later model parameters.
This is relevant for privacy-sensitive settings, by maintaining separate models for each task, making it ideal when data or models cannot be shared.
For example, for electronic health records, where hospitals face strict data-sharing restrictions or retention policies \citep{munoz2019}.
On-device learning is another key application, with domains such as domestic robots or monitoring cameras \citep{kapoor2021,verwimp2024continual}.

\textbf{Results:} \Cref{fig:smoother} illustrates the benefits of smoothing on \textsc{Gradual} \camelyon.
Smoothing passes information backwards, informing earlier models about the knowledge gained on later tasks, increasing performance
(\cref{fig:smoother} \textit{(left)}).
Across all $8$ seeds and tasks, we see a performance increase when comparing the filtered model to the smoothed model (\cref{fig:smoother} \textit{(right)}).
In particular, accuracy for the second task increases substantially from $70.5\%$ to $75.9\%$.
Smoothing may be particularly useful in sequences of tasks with only little data---and thus transferring knowledge between tasks is required---and the tasks are substantially different or challenging that a single model cannot accurately learn all tasks.

Additionally, Bayesian smoothing can also provide a grounded way to infer model parameters for novel tasks between two observed tasks.
Inferring a neural network's trained parameters for tasks for which we never directly observed any data, \eg for intermediate brightness levels, provides an interesting avenue for future work.

\subsection{Diagonal Plus Low-Rank Approximations}
\label{sec:exp_GGN}

Lastly, we analyze our proposed diagonal plus low-rank curvature approximation, which allows efficient computation of all filtering \& smoothing operations.
Here, we evaluate it outside of the filtering \& smoothing context, isolating the effectiveness of this approximation.
We compare it to common approximations from the continual learning literature.
The results show that adding a low-rank part---even with very small rank---adds useful information, while still offering the advantage of efficient filtering \& smoothing.
\begin{figure*}[!htb]
	\centering
	\includegraphics[width=0.95\linewidth]{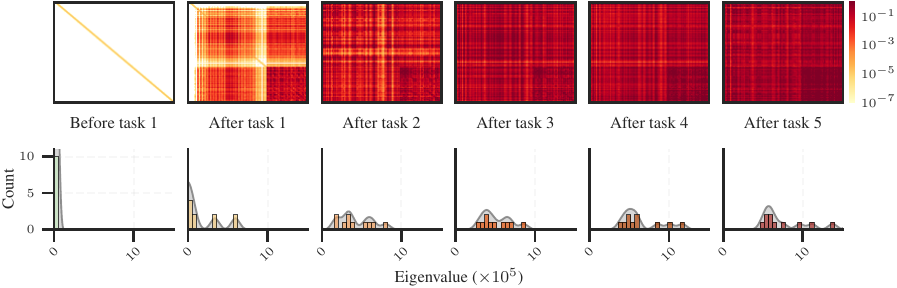}
	\caption{\textbf{The low-rank approximation across training.} \textit{(Top)} Our rank $k\!=\!10$ approximation to the precision matrix across tasks on \textit{Permuted} \mnist (see \cref{fig:eigs_20} for $k\!=\!20$). \textit{(Bottom)} Histograms of the eigenvalues of the approximation's low-rank part. With growing $t$, the eigenvalues increase in magnitude, indicating larger certainty and less flexibility.}
	\label{fig:eigs}
\end{figure*}
\begin{table}[!htb]
	\centering
	\caption{\textbf{Comparing our diagonal low-rank GGN approximation to other regularization methods}. We report the final \emph{average} accuracy ($\pm$ one standard deviation across $8$ seeds) on all sub-tasks after sequentially learning each task.}
	\label{tab:benchmark}
	\begin{tabular}{lll}
		\toprule
		& \textbf{\begin{tabular}[c]{@{}l@{}}\emph{Permuted}\\ \mnist\end{tabular}} & \textbf{\begin{tabular}[c]{@{}l@{}}\emph{Disjoint}\\ \mnist\end{tabular}} \\ \midrule
		\makecell[tl]{\textbf{LR-LGF}\\ our method ($\mQ\!=\!0$)}       & $0.964 \pm 0.004$ & $0.678 \pm 0.024$ \\
		EWC                       & $0.931 \pm 0.024$ & $0.676 \pm 0.029$ \\
		OSLA                      & $0.972 \pm 0.001$ & $0.852 \pm 0.004$ \\ \addlinespace
		\makecell[tl]{Baseline\\ (no regularizer)} & $0.664 \pm 0.020$ & $0.460 \pm 0.005$ \\ \bottomrule
	\end{tabular}
\end{table}

\textbf{Setting:}
We compare our diagonal plus low-rank approximation of the GGN to both EWC (where the regularizer differs as shown in \Cref{app:CL_approaches} and uses diagonal approximations of the Fisher matrix) and OSLA (which is the most similar but uses block-diagonal K-FAC approximations of the Hessian).
The comparison is performed on standard continual learning benchmarks.
In \emph{Permuted \mnist}, each task $t\!\in\!\{1,\dots,5\}$ consist of classifying the \mnist digits, where each task uses a random (but for this task fixed) permutation of the image pixels.
The \emph{Disjoint \mnist} setting consists of $2$ tasks: task $t\!=\!1$ includes images with labels $y\!\in\!\{1,\mathellipsis,4\}$, while $t\!=\!2$ contains labels $y\!\in\!\{5,\mathellipsis,9\}$.
For both continual learning problems, we train a small 2-layer MLP with $400$ hidden units per layer sequentially on all tasks (see \cref{app:exp_details_setup} for full details).
After each task, we record the current model's performance on all tasks seen so far.
For each problem and method, we tune the regularization strength $\lambda$ on a grid, maximizing the average accuracy.
Each experiment is repeated with 8 random seeds.
In these experiments, we deliberately disable the filter and smoother (\ie $\mQ\!=\!0$) to focus exclusively on the quality of the curvature approximation for continual learning.
This limits the experiments to the shared functionality of EWC, OSLA, and our method, intentionally disregarding the additional benefits of our approach.

\textbf{Results:}
\Cref{tab:benchmark,fig:benchmark} show that regularizing with a diagonal plus low-rank GGN approximation consistently reduces forgetting in both settings compared to no regularization.
Additionally, our method slightly outperforms EWC, suggesting that the low-rank component captures useful uncertainty information lost in the combination of purely diagonal approximations of EWC.
We also observe that OSLA performs significantly better, implying that, at least on these popular toy problems, interactions \emph{between} layers matter less.
Extending our framework to also support efficient filtering \& smoothing with these block-diagonal K-FAC approximations is a promising direction for future work.

\textbf{Properties of the diagonal plus low-rank approximation:}
We further analyze the behavior of our diagonal plus low-rank approximation during training.
\Cref{fig:eigs} visualizes its evolving eigenvalues after training each task.
For $\mQ\!=\!0$, the diagonal part remains constant throughout training (\cref{eq:lgf:low_rank_predict}).
First, the diagonal plus low-rank approximation is significantly more expressive than diagonal ones like EWC, with the low-rank component capturing useful parameter relationships.
Second, relatively low ranks and Hessian batch size are sufficient to achieve a meaningful yet cheap approximation.
Increasing the rank beyond $\approx\!10$ (the number of classes) or the batch size beyond $\approx\!8$ offers little additional benefit (\cref{fig:eigs_20,fig:ranks,fig:bs_hes}).
Lastly, the magnitudes of the eigenvalues grow as training progresses, reflecting increased certainty in parameter values.
Since the top eigenvalues are computed as part of the truncated SVD (line 7 in \cref{alg:low_rank_filtering}), they can serve as a cost-effective diagnostic.
For instance, they could indicate whether the network's ``memory'' has reached capacity, and the ratio of retained to deflated eigenvalues may inform the choice of the rank $k$.

\section{Conclusion}
\label{sec:conclusion}

\textbf{Limitations:}
Our diagonal plus low-rank GGN approximation stores matrices of size $1\!\times\!\numparams$, $\numparams\!\times\!\numclasses$, $\numclasses\!\times\!\numclasses$,
where $\numclasses$ is the model's output dimension.
This is beneficial if $\numclasses\!\ll\!\numparams$, but imposes constraints if the number of classes $\numclasses$ is large.
In such cases, other curvature approximations like K-FAC, block-diagonal or last-layer could be chosen.
However, these likely further reduce expressivity, and efficient filtering \& smoothing is not straightforward.
Our method may be infeasible for models with millions of parameters.
Furthermore, low-rank approximations with truncated SVDs can lead to an overestimated curvature, which can hurt performance.
While incorporating task-specific knowledge into the dynamics model is desirable, it can be challenging to set the associated parameters well in practical applications.
Finally, our experiments were designed to showcase specific scenarios serving as proxies for real-world continual learning challenges and to highlight our method's strengths.

\textbf{Summary and discussion:}
We have presented an efficient low-rank Laplace--Gaussian filtering framework for sequential deep learning across multiple related tasks.
It treats the network's weights as states, and the individual tasks as likelihood models in a Bayesian state-space model.
A diagonal plus low-rank Gaussian approximation allows for efficient approximate inference when combined with a low-rank Laplace approximation via the GGN matrix to compute the filtering distributions, \ie the posterior over the network's weights given all past and present tasks.
We leverage this formalism to compute \emph{task-specific} models via Bayesian smoothing, incorporating knowledge from later tasks into earlier models \emph{without renewed access to the data}.
Our methodology maps the complex problems of sequential or continual learning to well-understood Bayesian filtering and smoothing.
This addresses two key challenges of continual learning from \citet{verwimp2024continual}: It raises computational efficiency, and conceptually clarifies the method's components, casting the relations between datasets through the observation and dynamics models of Markov Chains.

%
\input{appendix/acknowledgments.tex}

%


\section*{Impact Statement}

The research in this paper is of foundational nature and its goal is to advance the field of Machine Learning. Ethical questions are thus not sufficiently prominent in this work to warrant a dedicated discussion section. In general, we believe that this work will have an overall positive impact as it can help improve the efficiency of neural network training.


\bibliographystyle{appendix/icml2025}
\bibliography{appendix/bibliography}


\newpage
\appendix
\onecolumn
\input{appendix/appendix.tex}

\end{document}

%% file: preamble/preamble.tex
\documentclass{article}
\PassOptionsToPackage{noend}{algorithmic}

\usepackage{amsfonts}				
\usepackage{amsmath}				
\usepackage{siunitx}  				
\usepackage{xfrac}					
\usepackage{bm}						
\usepackage{mathabx}				
\usepackage{mathtools}
\usepackage{amssymb}
\usepackage{amsthm}

\usepackage[english]{babel} 		

\usepackage{makecell}
\usepackage{hyphenat}

\usepackage{xspace}
\usepackage[utf8]{inputenc} 
\usepackage[T1]{fontenc}    
\usepackage{microtype}      
\usepackage{multicol}
\usepackage{enumitem}

\usepackage{caption}
\usepackage{subfigure}
\usepackage{tikz}
\usepackage{listofitems}
\usetikzlibrary{arrows,fit,backgrounds,positioning,shapes.multipart,shapes.geometric}
\tikzset{>=stealth'}
\usepackage{xcolor}
\usepackage{graphicx} 
\usepackage{placeins}

\usepackage{booktabs}               
\usepackage{multirow}               
\usepackage{soul}                   
\usepackage{changepage,threeparttable} 
\usepackage{tabularx}               
\usepackage{multirow}               
\usepackage{etoolbox}               
\usepackage{booktabs}
\usepackage{rotating}
\usepackage{array}               
\newrobustcmd{\B}{\bfseries}

\definecolor{mydarkblue}{rgb}{0,0.08,0.45} 
\usepackage[
	colorlinks=true,
	linkcolor=mydarkblue,
	citecolor=mydarkblue,
	filecolor=mydarkblue,
	urlcolor=mydarkblue,
]{hyperref}       		

\usepackage[nameinlink,capitalize]{cleveref}
\usepackage{url}            		

\usepackage[textsize=tiny]{todonotes}
\setuptodonotes{bordercolor=white}

\DeclareRobustCommand{\colordot}[1]{%
	\begin{tikzpicture}[baseline=(a.south)]
		\node[circle, scale=0.75,color=white, fill=#1] (a) {};
	\end{tikzpicture}%
}

\DeclareRobustCommand{\colorline}[1]{%
	\begin{tikzpicture}
		\raisebox{1.5pt}{
			\draw[#1,solid,line width=1.5pt] (0,0) -- (1em,0);
		}
	\end{tikzpicture}%
}

\DeclareRobustCommand{\colorarrow}[1]{%
	\begin{tikzpicture}
		\raisebox{1.5pt}{
			\draw[#1,dashed, line width=0.3mm, ->] (0,0) -- (1.5em,0);
		}
	\end{tikzpicture}%
}

\DeclareRobustCommand{\colorreversearrow}[1]{%
	\begin{tikzpicture}
		\raisebox{1.5pt}{
			\draw[#1,dashed,line width=0.3mm, <-] (0,0) -- (1.5em,0);
		}
	\end{tikzpicture}%
}

\newcommand{\colorcross}[1]{\textcolor{#1}{\ding{54}}}

\usepackage{pifont}
\usepackage[most]{tcolorbox}

%% file: preamble/commands.tex
\newcommand*{\eg}{e.g.\@\xspace}
\newcommand*{\ie}{i.e.\@\xspace}

\usepackage[super]{nth}						                

\newcommand*{\numrank}{k}

\newcommand{\dataset}{dataset\xspace}
\newcommand{\datasets}{datasets\xspace}

\newcommand{\UoT}{University of Tübingen}
\newcommand{\aicenter}{Tübingen AI Center}
\newcommand{\vectorinst}{Vector Institute}

\newcommand{\mnist}{\textsc{MNIST}\xspace}
\newcommand{\camelyon}{\textsc{CAMELYON}\xspace}



%% file: preamble/math_commands.tex
\DeclareMathOperator*{\argmin}{arg\,min}					
\newcommand{\bigO}{\mathcal{O}}								
\newcommand{\defeq}{\vcentcolon=}							
\newcommand{\eqdef}{=\vcentcolon}							

\renewcommand*{\mathellipsis}{%
	\mathinner{{\ldotp}{\ldotp}{\ldotp}}%
}															

\newcommand{\R}{\mathbb{R}}					
\newcommand{\N}{\mathbb{N}}					

\newcommand*{\numclasses}{C}
\newcommand*{\numtraindata}{N}
\newcommand*{\numparams}{D}

\newcommand*{\params}{\vtheta}

\newcommand*{\Loss}{\mathcal{L}}
\newcommand*{\RegLoss}{\mathcal{L}^\text{reg}}

\newcommand{\dd}{\,\text{d}}

\def\vtheta{{\bm{\theta}}}

\def\vg{{\bm{g}}}

\def\vm{{\bm{m}}}

\def\vx{{\bm{x}}}
\def\vy{{\bm{y}}}

\def\mA{{\bm{A}}}

\def\mC{{\bm{C}}}
\def\mD{{\bm{D}}}

\def\mG{{\bm{G}}}
\def\mH{{\bm{H}}}

\def\mJ{{\bm{J}}}

\def\mP{{\bm{P}}}
\def\mQ{{\bm{Q}}}

\def\mU{{\bm{U}}}
\def\mV{{\bm{V}}}
\def\mW{{\bm{W}}}

\def\mSigma{{\bm{\Sigma}}}

\DeclareMathAlphabet{\mathsfit}{\encodingdefault}{\sfdefault}{m}{sl}
\SetMathAlphabet{\mathsfit}{bold}{\encodingdefault}{\sfdefault}{bx}{n}

\def\sD{{\mathbb{D}}}

%% file: preamble/colors.tex
\definecolor{TUred}{RGB}{165,30,55}
\definecolor{TUdark}{RGB}{50,65,75}
\definecolor{TUgold}{RGB}{180,160,105}

\definecolor{TUmauve}{RGB}{180,160,150}

\definecolor{TUgray}{RGB}{185,184,188}

\definecolor{TUdarkblue}{RGB}{65,90,140}
\definecolor{TUblue}{RGB}{0,105,170}
\definecolor{TUlightblue}{RGB}{80,170,200}

\definecolor{TUlightgreen}{RGB}{125,165,75}
\definecolor{TUgreen}{RGB}{125,165,75}
\definecolor{TUdarkgreen}{RGB}{50,110,30}

\definecolor{TUlightred}{RGB}{200,80,60}
\definecolor{TUpurple}{RGB}{175,110,150}
\definecolor{TUorange}{RGB}{210,150,0}

\definecolor{statecolor}{HTML}{3D5F74}
\definecolor{measurementcolor}{HTML}{FFF6D4}

\definecolor{SNSblue}{rgb}{0.1216, 0.4666, 0.7059}
\definecolor{SNSorange}{rgb}{1.0, 0.4980, 0.0549}
\definecolor{SNSgreen}{rgb}{0.1725, 0.6274, 0.1725}
\definecolor{SNSred}{rgb}{0.84, 0.15, 0.16}
\definecolor{SNSpurple}{rgb}{0.58, 0.40, 0.74}

\definecolor{SNSorange_shaded}{HTML}{ffcea3}
\definecolor{SNSblue_shaded}{HTML}{8ebad9}
\definecolor{SNSgreen_shaded}{HTML}{cae7ca}
\definecolor{SNSred_shaded}{HTML}{ea9293}

\definecolor{MPLred_shaded}{HTML}{df735b}
\definecolor{MPLblue_shaded}{HTML}{3885bc}

\definecolor{statecolor}{HTML}{3D5F74}
\definecolor{measurementcolor}{HTML}{FFF6D4}
\definecolor{datasetcolor}{HTML}{FFF6D4}
\definecolor{darkgrey}{HTML}{6C6C6C}

%% file: figs/visual_abstract.tex
\begin{tikzpicture}[
	params/.style={circle, draw=none, minimum size=30pt, fill=statecolor!50},
	datasets/.style={circle, draw=none, minimum size=30pt, fill=TUgold!50},
	black_arrow/.style={->, semithick},
	smoothing_arrow/.style={<-, dashed, line width=0.35mm, TUlightred, bend angle=10, bend left, opacity=0.7},
	predict_arrow/.style={->, dashed,  line width=0.35mm, TUlightblue, bend angle=10, bend right},
	update_arrow/.style={->, dashed, line width=0.35mm,  TUpurple, bend angle=55, bend left},
	NN_node/.style={circle, draw=none, minimum size=0.25pt, fill=TUlightblue},
	]
	
	\node[params] (tm) at (0,0) {$\params_{t-1}$};
	\node[params] (tt) at (5,0) {$\params_{t}$};
	\node[params] (tp) at (10,0) {$\params_{t+1}$};
	
	\node[] (ldots) at (-2,0) {$\dots$};
	\node[] (rdots) at (12,0) {$\dots$};
	\draw[black_arrow] (ldots) edge (tm);
	\draw[black_arrow] (tp) edge (rdots);
	
	\node[datasets] (dm) at (0,-3.5) {$\sD_{t-1}$};
	\node[datasets] (dt) at (5,-3.5) {$\sD_{t}$};
	\node[datasets] (dp) at (10,-3.5) {$\sD_{t+1}$};
	
	\draw[black_arrow] (tm) edge (tt);
	\draw[black_arrow] (tt) edge (tp);
	\draw[black_arrow] (tm) edge (dm);
	\draw[black_arrow] (tt) edge (dt);
	\draw[black_arrow] (tp) edge (dp);
	
	\foreach \s in {0,...,2}{
		\begin{scope}[xshift=\s * 5cm, yshift=-1.0cm, scale=0.45, rotate=-90] 
			\readlist\Nnod{5,3,2} 
			\foreachitem \N \in \Nnod{
				\foreach \i [evaluate={\x=\Ncnt; \y=\N/2-\i+0.5; \prev=int(\Ncnt-1);}] in {1,...,\N}{
					\node[NN_node] (N\Ncnt-\i) at (\x,\y) {};
					\ifnum\Ncnt>1
					\foreach \j in {1,...,\Nnod[\prev]}{
						\draw[] (N\prev-\j) -- (N\Ncnt-\i);
					}
					\fi
				}
			}
		\end{scope}
	}
	
	\draw[smoothing_arrow] (tt.north east) to node[below] {\textbf{Smoothing}} (tp.north west);
	\draw[smoothing_arrow] (tm.north east) to (tt.north west);
	
	\draw[predict_arrow] (tm.south east) to (tt.south west);
	\draw[predict_arrow] (tt.south east) to node [below] {
		\begin{minipage}{3cm}
			\centering
			\textbf{Predict} \\[-0.2em]
			$\textcolor{black}{p(\params_{t+1}\mid \params_{t})}$ \\
			$\textcolor{black}{=\!\mathcal{N}(\params_{t+1}; \params_t, \textcolor{TUlightblue}{\mQ_t})}$
		\end{minipage}
	} (tp.south west);
	
	\draw[update_arrow] (dm.north west) to (tm.south west);
	\draw[update_arrow] (dt.north west) to node [left, xshift=0.45cm, yshift=-0.58cm] {
		\begin{minipage}{3.5cm}
			\centering
			\textbf{Train \& Update} \\[-0.2em]
			$\textcolor{black}{p(\params_{t}\mid \sD_t)}$ \\
			$\textcolor{black}{\textcolor{TUpurple}{\approx} \mathcal{N}(\params_t;\params^{\ast}_{t},\mC_t)}$\\
			\textbf{Laplace}\\[-0.2em]
			\textbf{Approximation}\\[-0.2em]
			\ \\[-0.6em]
			\textcolor{black}{$\params^{\ast}_{t}$ trained with}\\
			\textbf{diagonal plus low-rank}\\[-0.2em]
			\textbf{regularizer} $\mC_{t-1}$
		\end{minipage}
	} (tt.south west);
	\draw[update_arrow] (dp.north west) to (tp.south west);

\end{tikzpicture}

%% file: figs/algorithm.tex
\newtcolorbox{predictblock}{
	boxrule=0pt,
	colback=TUlightblue!15,
	left=0em,
	left skip=-0.3em,
	before skip=0.5em,
	after skip=0em,
	right=0em,
	top=0em,
	bottom=0em,
	leftrule=0pt,
	colframe=TUlightblue!50
}

\newtcolorbox{updateblock}{
	boxrule=0pt,
	colback=TUpurple!15,
	left=0em,
	left skip=-0.3em,
	before skip=0.5em,
	after skip=0.5em,
	right=0em,
	top=0em,
	bottom=0em,
	leftrule=0pt,
	colframe=TUpurple!50
}

\newcommand{\RETURNE}[1]{\hspace{-0.6cm} \textbf{Return:} #1}
\newcommand{\codecomment}[1]{{\color{darkgray!75} {\slash\slash\,}\itshape #1}}

\begin{algorithm}[tb]
	\caption{Low-rank Laplace--Gaussian Filter (LR-LGF)}
	\label{alg:low_rank_filtering}
	\begin{algorithmic}[1]
		\REQUIRE
		Loss functions \(\Loss_t\) for tasks \(t\!=\!1,\!\mathellipsis\!,T\)\!,
		initial mean \(\vm_0\) and precision \(\mP_0 \!=\! \mD_0 \!+\! \mU_0 \mSigma_0 \mU_0^\top\!\),
		process noise covariance \(\mQ\),
		regularization strength \(\lambda\),
		rank \(\numrank\).
		\FOR{\(t = 1\) \TO \(T\)}
		\begin{predictblock}
			\STATE \textbf{\color{TUblue!90}Predict}\\
				\codecomment{Compute the predicted mean \(\vm_t^-\) and precision \(\mP_t^- \!=\! \mD_t^- \!+\! \mU_t^- \mSigma_t^- \mU_t^{-\top}\)}
				\STATE \(\vm_t^-\!, \mD_t^-\!, \mU_t^- \!\leftarrow\! \vm_{t-1}, \left( \mQ + \mD_{t-1}^{-1} \right)^{-1}\!\!\!\!, \mD_{t-1}^{-1} \mU_{t-1}\)
		\end{predictblock}
		\begin{updateblock}
			\STATE \textbf{\color{TUpurple!90!black}Update}\\
				\codecomment{Train on task \(t\) with the regularized loss \(\RegLoss_t\)}
				\STATE \(\vtheta_t^* \leftarrow \argmin_{\vtheta} \RegLoss_t(\vtheta)\)\\
				\codecomment{Compute mini-batched GGN \(\sum_{b=1}^B \!\mJ_t^{(b)}\! \hat{\mH}_t^{(b)\!} {\mJ_t^{(b)}}^{\!\top}\)}
				\STATE  \(\mJ_t^{(b)}, \hat{\mH}_t^{(b)} \leftarrow (\frac{\partial f}{\partial \params})^{(b)}, (\frac{\partial^2 \Loss}{\partial f^2})^{(b)} \quad \text{for}\ b = 1, \mathellipsis, B.\)\\
				\codecomment{Perform a truncated SVD (see \cref{eq:W_t}).}
				\STATE \(\tilde{\mU}_t, \tilde{\mSigma}_t, \tilde{\mV}_t^\top \leftarrow \operatorname{tSVD}_{\numrank} \left(
\mW_t
							\right)\)\\
				\codecomment{Compute the filtering mean \(\vm_t\) and precision \(\mP_t = \mD_t + \mU_t \mSigma_t \mU_t^\top\)}
				\STATE \(\vm_t,\ \mD_t,\ \mU_t,\ \mSigma_t \leftarrow \vtheta_t^*,\ \mD_t^-,\ \tilde{\mU}_t,\ \tilde{\mSigma}_t^2\)
		\end{updateblock}
		\ENDFOR
		\RETURNE 
		Filtering means \((\vm_t)_{t=1}^T\) and diagonal plus low-rank precisions \( \left( \mD_t, \mU_t, \mSigma_t \right)_{t=1}^T \).
	\end{algorithmic}
\end{algorithm}

%% file: appendix/acknowledgments.tex

\section*{Acknowledgments}

The authors gratefully acknowledge co-funding by the Carl Zeiss Foundation, (project "Certification and Foundations of Safe Machine Learning Systems in Healthcare") and the European Union (ERC, ANUBIS, 101123955). Views and opinions expressed are however those of the author(s) only and do not necessarily reflect those of the European Union or the European Research Council. Neither the European Union nor the granting authority can be held responsible for them). Philipp Hennig is a member of the Machine Learning Cluster of Excellence, funded by the Deutsche Forschungsgemeinschaft (DFG, German Research Foundation) under Germany's Excellence Strategy – EXC number 2064/1 – Project number 390727645; he also gratefully acknowledges funds from the Ministry of Science, Research and Arts of the State of Baden-Württemberg.
Frank Schneider is supported by funds from the Cyber Valley Research Fund. 	
Joanna Sliwa and Nathanael Bosch are grateful to the International Max Planck Research School for Intelligent Systems (IMPRS-IS) for support.

%% file: appendix/appendix.tex

\section{Continual Learning}
\label{app:CL_definition}

\citet{Kirkpatrick2017} define continual learning as an ability to learn tasks sequentially and maintain knowledge for tasks from the past which are not experienced anymore. 
The aim is to have an agent that performs well across multiple tasks and can incorporate new information \citep{Ritter2018}. 
Our goal is to minimize the training loss summed over all tasks with the constraint that we can only access the loss of the current tasks \citep{zenke2017continual}. 
\citet{titsias2020functional} mention that we should not need an extensive retraining on previous maintained data. 
Recent review on continual learning \citep{review_cl} summarizes it as observing tasks sequentially but behaving as if seeing them simultaneously. We need to obtain a balance between learning flexibility and memory stability and generalization within and between tasks.
The book of lifelong learning \citep{book} thoroughly defines continual learning and its connections to transfer, multi-task, online and reinforcement learning. The authors define that at a given point in time, a model has learned a sequence of previous tasks $t\!=\!t_1, t_2, \dots, t_T$ with corresponding data $\sD\!=\!\{\sD_1, \sD_2, \dots, \sD_T\}$.  The tasks can be of the same or different type/domain. 
When the model encounters a new task $t_{T+1}$ with data $\sD_{T+1}$ it can leverage the past knowledge and learn the new task. Then it implements the new knowledge into the existing knowledge base. The main requirements are that the learning is in a continuous fashion, the knowledge is accumulated and the knowledge from the previous tasks can be used to learn new ones.

\subsection{Additional Approaches to Continual Learning}
\label{app:CL_approaches}

In \Cref{sec:related_work}, we focus on regularization-based approaches to continual learning, as this is the approach used in the paper. 
Below, we present the summary of the used competing methods:

\textbf{EWC:} Based on the descriptions in the original paper \citep{Kirkpatrick2017}, as well as, mentions in subsequent papers \citep{EWC-indepth, EWC-mention} we use the following loss function when learning task $t$:
$$\mathcal{L}(\theta) = \mathcal{L}_t (\theta) +\frac{\lambda}{2} \sum_k^{T-1} (\theta-\theta^{*}_{k})^T\hat{F}_{k}(\theta-\theta^{*}_{k})$$
$T$ is the number of all tasks, $\theta^{*}$ is the mode for given task, $\hat{F}$ is the diagonal of the Fisher Information Matrix of the negative log-likelihood $F$, given by
$$F=\frac{1}{|\mathcal{S}_k|}\sum^{\mathcal{S}_k}_{s_k} \left(-\frac{\delta \log p({s}_k|\theta)}{\delta \theta}\bigg|_{\theta=\theta_k^*}\right)^2$$ where $\mathcal{S}_k$ is a subset of the data.

\textbf{OSLA:} For the second method we use the following loss funtion when learning task $t$ based on the original paper \citep{Ritter2018}:
$$\mathcal{L}(\theta) = \mathcal{L}_t(\theta) +\frac{1}{2}(\theta -\theta^{*}_{t-1})^T\Lambda_{t-1}(\theta_t -\theta^{*}_{t-1})$$
where $\Lambda_{t-1}$ is sum of the scaled Hessian of the negative log-likelihoods (the Hessians are computed using KFAC): $$\Lambda_{t-1} = \lambda \left(-\frac{\delta^2 \log p(\mathcal{D}_{t-1}|\theta)}{\delta \theta\delta \theta}\bigg|_{\theta=\theta^{*}_{t-1}}\right) + \Lambda_{t-2} $$
For completeness, we also briefly summarize alternative approaches to continual learning below, following the taxonomy by \citet{review_cl}.

\textbf{Replay-based:} Relay-based methods' key idea, mainly of experience replay, is to store a small set of training data points from the past tasks. Methods that use this approach are \eg \citep{klasson2022learn,rolnick2019experience}.
One of the challenges is selecting representative samples and ensuring the memory buffer storage is efficient.

\textbf{Optimization-based:} These methods are based on changing the optimization design, \eg by allowing gradient updates only in the orthogonal directions to previous tasks or finding only flat local minima.

\textbf{Representation-based:} These methods create and use representations for each tasks through self-supervised learning or pre-training.

\textbf{Architecture-based:} Researchers have also tried parameter allocation, model decomposition or modular networks.
They either isolate some parameter subspace for a certain tasks, separate the model into task-sharing and task-specific components, or divide the network into modules dedicated to special tasks.

\textbf{Gaussian Processes for CL:}
The state-space model that we consider in this work has a continuous-time representation as a Gaussian process with a Wiener process kernel.
The GP maps the task-id to the neural network weights, that is $(t \mapsto \theta_t)$, but the actual model of interest which maps data inputs to outputs, i.e. $(x \mapsto y)$, is the neural network inside the observation model.
In the related works \citep{munoz2019, kapoor2021}, the GP is used to learn the function of interest $(x \mapsto y)$. There is no fundamental distinction between tasks and thus there is no prior which encodes how continuous or distinct weights between those should be, which is precisely the role of the GP in our work. 
\citet{Titsias2020} combines GPs and NNs and considers GPs with deep kernel functions, but here again the GP is an integral part of the function of interest as it outputs the quantity of interest $y$. 
In contrast, the GP in our work plays a much more subtle role and only provides a prior over the neural network weights varying per task, and it is the neural network that plays the more prominent role and maps the data inputs to outputs.

\textbf{CL vs multi-task learning:}
\citet{mirzadeh2020} point out that when the model has access to all the data, it learns different solutions than when learning continuously.

\textbf{Task similarity:}
The similarity of tasks in continual learning was studied in previous works \citep{adel2024, ramesh2022, wang2022}. 
Importantly, the authors state that one may benefit from the knowledge of task relatedness. 
In such a setting, we can employ an ensemble of small models that grows when competing (vs synergistic) tasks appear. 
When dissimilar tasks are learnt by one model, the tasks compete for the fixed capacity of the model. 
Additionally, evaluating the task similarity and initializing some components with the most similar past task’ one may help with the training.

\section{Implementation Details}

\label{sec:implementation}

We implemented a class object \textsc{Diag\_LowRank} has the following methods:
	\begin{itemize}
	\item $\textsc{\_\_init\_\_}$ -- creating a diagonal matrix scaled by $\lambda_{\textrm{INIT}}$ and a zero low-rank term,
	\item $\textsc{add\_compute\_inv\_sum}$ -- inverting the precision according to the Woodbury identity, adding a diagonal matrix and final inverting 
	\item $\textsc{add\_low}$ -- adding the computed precision to the stored diagonal plus low-rank term (uses square roots), and deflating down to a memory limit (which the user provides as an optional argument, otherwise the deflation to $C$) via truncated SVD,
	\item $\textsc{update\_mP}$ -- computing the vector matrix product of mean and the diagonal plus low rank precision,
	\item $\textsc{smooth}$ -- updating the mean based on the smoothing gain.
  \end{itemize}
\section{Filtering and Smoothing with Diagonal plus Low-Rank Matrices}

\subsection{Mathematical operations on diagonal plus low-rank matrices}
\label{app:low_rank_math}
In the following we describe how to perform mathematical operations on diagonal plus low-rank matrices \(\mP = \mD + \mU \mSigma \mU^\top\), where
\(\mD \in \R^{\numparams \times \numparams}\) is a diagonal matrix,
\(\mU \in \R^{\numparams \times \numrank}\) is a tall matrix,
and \(\mSigma \in \R^{\numrank \times \numrank}\) is a dense matrix,
with \(\numrank \ll \numparams\).
And importantly, all of these operations should be such that they maintain the low-rank structure of the matrix such that we never need to store the full matrix in memory,
and they should be computationally efficient and scale at most linearly in \(\numparams\).
\begin{itemize}[left=\parindent]
\item
  \textbf{Addition with a diagonal matrix:}
  Adding a diagonal matrix to a diagonal plus low-rank matrix results in a diagonal plus low-rank matrix:
  \begin{equation}
    \left( \mD + \mU \mSigma \mU^\top \right) + \Lambda
    = \underbracket{\left( \mD + \Lambda \right)}_{\mD'} + \mU \mSigma \mU^\top
  \end{equation}
\item
  \textbf{Matrix inversion:}
  The inverse of a diagonal plus low-rank matrix can be computed efficiently using the Woodbury matrix identity:
  \begin{equation}
    \left( \mD + \mU \mSigma \mU^\top \right)^{-1}
    =
    \underbracket{\mD^{-1}}_{\mD'} +
    \underbracket{\mD^{-1} \mU}_{\mU'}
    \underbracket{\left( - \left(  \mSigma^{-1} - \mU^\top \mD^{-1} \mU \right)^{-1} \right)}_{\mSigma'}
    \underbracket{\mU^\top \mD^{-1}}_{\left( \mU' \right)^\top}
  \end{equation}
\item
  \textbf{Addition of two low-rank matrices:}
  Adding two diagonal plus low-rank matrices results in a diagonal plus low-rank matrix, but with increased rank \(\numrank' = \numrank_1 + \numrank_2\):
  \begin{equation}
    \left( \mD_1 + \mU_1 \mSigma_1 \mU_1^\top \right) + \left( \mD_2 + \mU_2 \mSigma_2 \mU_2^\top \right)
    = \left( \mD_1 + \mD_2 \right) +
    \begin{bmatrix} \mU_1 & \mU_2 \end{bmatrix}
    \begin{bmatrix} \mSigma_1 & 0 \\ 0 & \mSigma_2 \end{bmatrix}
    \begin{bmatrix} \mU_1^\top \\ \mU_2^\top \end{bmatrix}
  \end{equation}
  Alternatively, to keep the rank low we can perform a truncated singular value decomposition on the matrix square-root:
  \begin{equation}
    \tilde{\mU} \tilde{\mSigma} \tilde{\mV}^\top
    = \operatorname{tSVD}_\numrank \left( \begin{bmatrix} \mU_1 \mSigma_1^{1/2} & \mU_2 \mSigma_2^{1/2} \end{bmatrix} \right)
  \end{equation}
  and then approximate
  \begin{equation}
    \left( \mD_1 + \mU_1 \mSigma_1 \mU_1^\top \right) + \left( \mD_2 + \mU_2 \mSigma_2 \mU_2^\top \right)
    \approx \underbracket{\left( \mD_1 + \mD_2 \right)}_{\mD'} +
    \underbracket{\tilde{\mU}}_{\mU'}
    \underbracket{\tilde{\mSigma}^2}_{\mSigma'}
    \underbracket{\tilde{\mU}^\top}_{\left( \mU' \right)^\top}
  \end{equation}
\end{itemize}

\subsection{Low-Rank Kalman Predict Step in Information Form}
\label{app:low_rank_predict}
Given a precision matrix \(P\) and a transition noise covariance \(Q\), the predict step in a Gaussian filter computes the predictive precision as
\begin{equation}
  \mP^- = \left( \mP^{-1} + \mQ \right)^{-1}.
\end{equation}
Now if the precision is
a diagonal plus low-rank matrix \(\mP = \mD + \mU \mSigma \mU^\top\),
and the transition noise covariance \(\mQ\) is diagonal,
we can show that the predictive precision is also diagonal plus low-rank.
First, we apply the Woodbury matrix identity to the precision matrix:
\begin{align}
  \left( \mQ + \mP^{-1} \right)^{-1}
  &= \left( \mQ + \left(\mD + \mU \mSigma \mU^\top \right)^{-1} \right)^{-1} \\
  &= \left( \mQ + \mD^{-1} - \mD^{-1} \mU \left( \mSigma^{-1} + \mU^\top \mD^{-1} \mU \right)^{-1} \mU^\top \mD^{-1} \right)^{-1}.
\end{align}
Defining
\(\mD':= \mQ+\mD^{-1}\),
\(\mU':= \mD^{-1} \mU\),
and \(\mSigma':= -\left( \mSigma^{-1} + \mU^\top \mD^{-1} \mU \right)^{-1}\),
and applying the Woodbury matrix identity again, we get
\begin{align}
  \left( \mQ + \mP^{-1} \right)^{-1} = \mD'^{-1} - \mD'^{-1} \mU' \left( \mSigma'^{-1} + \mU'^\top \mD'^{-1} \mU' \right)^{-1} \mU'^\top \mD'^{-1}
\end{align}
This shows that the predictive precision is also diagonal plus low-rank:
\begin{equation}
  \mP^- = \mD^- + \mU^- \mSigma^- (\mU^-)^\top,
\end{equation}
with
\begin{align}
  \mD^- &:= \mD'^{-1} \\
       &\ = \left( \mQ + \mD^{-1} \right)^{-1} \\
  \mU^- &:= \mD'^{-1} \mU' \\
       &\ = \left( \mQ + \mD^{-1} \right)^{-1} \mD^{-1} \mU, \\
  \mSigma^- &:= -\left( \mSigma'^{-1} + \mU'^\top \mD'^{-1} \mU' \right)^{-1} \\
       &\ = \left(
         \mSigma^{-1} + \mU^\top \mD^{-1} \mU
         - \mU^\top \mD^{-1} \left( \mQ  + \mD^{-1} \right)^{-1} \mD^{-1} \mU
         \right)^{-1}
\end{align}
This, to perform the Kalman predict step in information form with a diagonal plus low-rank precision matrix, we compute and return the above quantities
\(\mD^-\), \(\mU^-\), and \(\mSigma^-\).

\subsection{Low-Rank Kalman Smoother Step in Information Form}
\label{app:low_rank_smoother}
\label{app:low_rank_update}

The standard Kalman smoother, or Rauch--Tung--Striebel smoother \citep{rauchtungstriebel1965},
computes Gaussian posterior distributions
\begin{equation}
  p(\params_{t} \mid \sD_{1:T}) = \mathcal{N}(\params_t;  \vm_t^s, \mC_t^s)
\end{equation}
by iterating the following backward recursion, starting with the filtering distribution
\(\vm_T^s = \vm_T\) and \(\mC_T^s = \mC_T\):
\begin{align}
  \mG_t &= \mC_t \left( \mC_t^- \right)^{-1}, \\
  \label{app:eq:smoother_mean}
  \vm_t^s &= \vm_t + \mG_t \left( \vm_{t+1}^s - \vm_{t+1}^- \right), \\
  \mC_t^s &= \mC_t + \mG_t \left( \mC_{t+1}^s - \mC_{t+1}^- \right) \mG_t^\top.
\end{align}
Now let us formulate the smoother step in terms of diagonal plus low-rank precision matrices.
Let
the filtering precision at time \(t\) be
\(\mP_t = \mD_t + \mU_t \mSigma_t \mU_t^\top\),
the smoothing precision at time \(t+1\) be
\(\mP_{t+1}^s = \mD_{t+1}^s + \mU_{t+1}^s \mSigma_{t+1}^s (\mU_{t+1}^s)^\top\),
and let \(\mQ\) be diagonal.
Recall that the predicted precision satisfies
\(\mP_{t+1}^- = \left( \mP_{t}^{-1} + \mQ \right)^{-1}\).
Then, the smoothing gain \(\mG_t\) is given by
\begin{equation}
  \mG_t
  = \mC_t \left( \mC_t^- \right)^{-1}
  = \mP_t^{-1} \mP_{t+1}^-
  = \mP_t^{-1} \left( \mP_t^{-1} + \mQ \right)^{-1}
  = \left( I + \mQ \mP_t \right)^{-1}.
\end{equation}
Plugging in the diagonal plus low-rank form of the precision matrices, we get
\begin{equation}
  \mG_t
  = \big(
    \underbracket{I + \mQ \mD_t}_{\mD'}
    +
    \underbracket{\mQ \mU_t}_{\mU'}
    \underbracket{\mSigma_t}_{\mSigma'}
    \underbracket{\mU_t^\top}_{\mV'^\top}
  \big)^{-1}
\end{equation}
Applying the Woodbury matrix identity, we get
\begin{equation}
  \mG_t
  = \mD'^{-1} - \mD'^{-1} \mU' \left( \mSigma'^{-1} + \mV'^\top \mD'^{-1} \mU' \right)^{-1} \mV'^\top \mD'^{-1}.
\end{equation}
Therefore, the smoothing gain is diagonal plus low-rank
\(\mG_t = \mD_t^G + \mU_t^G \mSigma_t^G (\mV_t^G)^\top\),
with
\begin{align}
  \mD_t^G &:= \mD'^{-1} \\
        &\ = \left( I + \mQ \mD_t \right)^{-1}, \\
  \mU_t^G &:= \mD'^{-1} \mU' \\
        &\ = \left( I + \mQ \mD_t \right)^{-1} \mQ \mU_t, \\
  \mSigma_t^G &:= - \left( \mSigma'^{-1} + \mV'^\top \mD'^{-1} \mU' \right)^{-1} \\
        &\ = -\left( \mSigma_t^{-1} + \mU_t^\top \left( I + \mQ \mD_t \right)^{-1} \mQ \mU_t \right)^{-1}, \\
  \left( \mV_t^G \right)^\top &:= \mV'^\top \mD'^{-1} \\
        &\ = \mU_{t}^\top \left( I + \mQ \mD_t \right)^{-1}.
\end{align}
This concludes the first part:
Computing the smoothed mean as in \Cref{app:eq:smoother_mean} can be done efficiently as \(\mG_t\) is diagonal plus low-rank.

The smoothing covariance/precision can again be approximated efficiently in a diagonal plus low-rank manner.
We first re-write the smoothing covariance in terms of precisions, and re-order some terms to obtain an addition of two low-rank matrices:
\begin{align}
  \left(\mP_t^s\right)^{-1}
  &= \left( \mP_t \right)^{-1} + \mG_t \left( \left( \mP_{t+1}^s \right)^{-1} - \left( \mP_{t+1}^- \right)^{-1} \right) \mG_t^\top \\
  &= \left( \mP_t \right)^{-1} + \mG_t \left( \left( \mP_{t+1}^s \right)^{-1} - \left( \mP_{t} \right)^{-1} - Q\right) \mG_t^\top \\
  &= \left( \mP_t \right)^{-1} - \mG_t \left( \mP_{t} \right)^{-1} \mG_t^\top + \mG_t \left( \left( \mP_{t+1}^s \right)^{-1} - Q\right) \mG_t^\top \\
  &= \left( I - \mG_t \right) \left( \mP_t \right)^{-1} \left( I - \mG_t \right)^\top + \mG_t \left( \left( \mP_{t+1}^s \right)^{-1} - Q\right) \mG_t^\top.
\end{align}
Then, since \(\mP_t\) and \(\mP_{t+1}^s\) are diagonal plus low-rank, their inverse is also diagonal plus low-rank (see \cref{app:low_rank_math}).
Let
\((\mP_t)^{-1} = \mD_t + \mU_t \mSigma_t \mU_t^\top\)
and
\((\mP_{t+1}^s)^{-1} = \mD_{t+1}^s + \mU_{t+1}^s \mSigma_{t+1}^s (\mU_{t+1}^s)^\top\).
Then, the smoothing precision can be written as
\begin{align}
  \left(\mP_t^s\right)^{-1}
  &= \left( I - \mG_t \right) \left( \mD_t + \mU_t \mSigma_t \mU_t^\top \right) \left( I - \mG_t \right)^\top \\
  &\quad + \mG_t \left( \mD_{t+1}^s + \mU_{t+1}^s \mSigma_{t+1}^s (\mU_{t+1}^s)^\top - \mQ \right) \mG_t^\top.
\end{align}
Similarly to before in \cref{app:low_rank_math}, we can write the smoothing precision as a matrix product \(\left(\mP_t^s\right)^{-1} = \mW \mW^\top\),
by defining \(\mW\) as
\begin{equation}
  \mW :=
  \begin{bmatrix}
    (I-\mG_t) \mD_t^{1/2}
    & (I-\mG_t) \mU_t \mSigma_t^{1/2}
    & \mG_t (\mD_{t+1}^s - \mQ)^{1/2}
    & \mG_t \mU_{t+1}^s (\mSigma_{t+1}^s)^{1/2}
  \end{bmatrix}.
\end{equation}
Then, we can perform a truncated SVD on \(\mW \approx \mU \mSigma \mV^\top\) to obtain a low-rank approximation of the smoothing covariance
\begin{equation}
  \left(\mP_t^s\right)^{-1} \approx \mU \mSigma^2 \mU^\top.
\end{equation}
A low-rank approximation of the precision follows again with the Woodbury matrix identity (see \cref{app:low_rank_math}).

\section{Experimental Details}
\label{app:exp_details}

\subsection{Datasets}
The datasets that we used are \textrm{MNIST} \citep{deng2012mnist} and \textrm{Camelyon} \citep{koh2021wilds}. Our implementation is in JAX \citep{jax2018github}.

\label{app:exp_details_data}

In Section \ref{sec:experiments} we use \textit{Permuted} and \textit{Disjoint} \mnist.
Next, we create three new settings within the \textsc{\camelyon} dataset from the WILDS benchmark, a collection of datasets designed to address distribution shifts commonly encountered in real-world scenarios. 
These include domain generalization, where the objective is to generalize to unseen domains during training, and subpopulation shift, where class proportions differ.
We adapted \textsc{\camelyon} for a continual learning setup, expanding it with additional examples that are simple yet illustrative of dataset shifts, following as a guideline the work by \citet{datashift_book}.
We believe they capture realistic scenarios e.g. brightness shifts due to different generations of machines, varying staining methods or imbalanced amount of data.
As the \textsc{gradual \camelyon} was already described in section \ref{sec:exp_Q}, here we only offer extended description and introduce the remaining two datasets.




\begin{figure}
  \centering
  \includegraphics[width=0.8\linewidth]{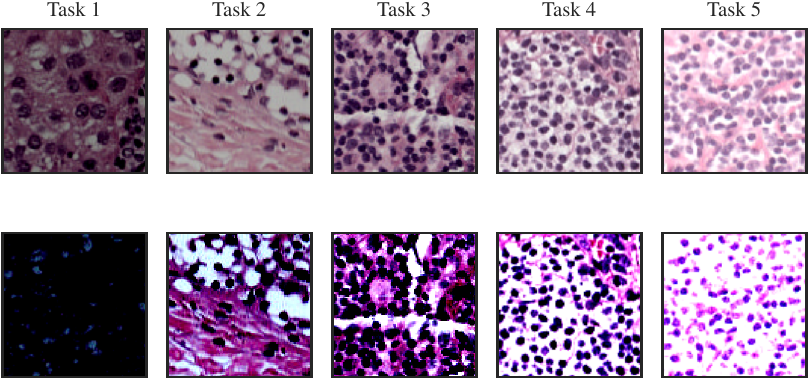}
  \caption{\textbf{Examples of the input data from} \textsc{Gradual CAMELYON}. (\textit{Top}) To adjust the brightness, we apply a shift $x_t = x + \Delta_t$ (\textit{bottom}) next, we normalize $(x_t - \mu_X)/\sigma_X$.}
  \label{fig:camelyon_images}
\end{figure}



\subsection{Experimental Setup}
\label{app:exp_details_setup}

Table \ref{tab:expsetup} reports the values of all hyperparameters for each result presented in the paper. 
The number of seeds controlled all the variability i.e. initialization, data loaders, dataset splits, random operations etc.
All of the experiments were run on local desktop with one NVIDIA GeForce RTX 2080 Ti with 11 GB memory or local laptop, Apple M3 Max, with 64 GB memory. The runtime of each experiment ranged from 2 minutes to 150 minutes.

\begin{table}
  \centering
  \caption{\textbf{Experimental setup: MNIST}. The hyperparameters were tuned via an independent search on a grid on validation set.}
  \label{tab:expsetup}
  \begin{tabular}{lrr}
      \toprule
      \textbf{Experiment} & \textbf{\begin{tabular}[c]{@{}c@{}}\textit{Permuted} \\ MNIST\end{tabular}} 
      & \textbf{\begin{tabular}[c]{@{}c@{}}\textit{Disjoint} \\ MNIST\end{tabular}} \\
      \midrule
      layers & \multicolumn{2}{c}{2} \\
      units & \multicolumn{2}{c}{400} \\
      epochs & \multicolumn{2}{c}{10} \\
      points & 60 000 & 12 000 \\
      $\lambda$ & \begin{tabular}[c]{@{}r@{}}ours: $10^5$, diag: $10^{-4}$\\ EWC: $10^9$\\ OSLA: 10\end{tabular} & \begin{tabular}[c]{@{}r@{}}ours: $5 \cdot 10^4$\\ EWC: $9 \cdot 10^5$\\ OSLA: 10\end{tabular} \\
      weight decay & 0 & $10^{-5}$ \\
      seeds & \multicolumn{2}{c}{8} \\
      batch size & \multicolumn{2}{c}{\begin{tabular}[c]{@{}c@{}}128\\ 128\end{tabular}} \\
      Hessian batch size & \multicolumn{2}{c}{128} \\
      learning rate & 0.001 & 0.00001 \\
      initial $\lambda$ & 0.0001 & 0.00001 \\
      \bottomrule
  \end{tabular}
\end{table}

\begin{table}
  \centering
  \caption{\textbf{Experimental setup: \textsc{CAMELYON}}. The hyperparameters were tuned via an independent search on a grid on validation set.}
  \label{tab:expsetup_part2}
  \begin{tabular}{lrr}
      \toprule
      \textbf{Experiment} 
      & \textbf{\begin{tabular}[c]{@{}c@{}}\textsc{Gradual} \\ \textsc{CAMELYON}\\ study of Q\\ (\Cref{sec:exp_Q})\end{tabular}}
      & \textbf{\begin{tabular}[c]{@{}c@{}}\textsc{Gradual} \\ \textsc{CAMELYON}\\ smoother\\ (\Cref{sec:exp_smoother})\end{tabular}} \\
      \midrule
      layers & {\begin{tabular}[c]{@{}c@{}}3 Conv\\ 2 Dense\end{tabular}} \\
      units & {\begin{tabular}[c]{@{}c@{}}32, (3x3)\\ 16, (3x3)\\ 4, (3x3)\\ 8, 2\end{tabular}} \\
      epochs & {c}{5} \\
      points & 5000 & 750 \\
      $\lambda$ &  $10^5$ & $10^4$ \\
      weight decay & {0} \\
      seeds & {8} \\
      $\mQ$ & \begin{tabular}[c]{@{}c@{}}structured: $10^{-5}$ \\(C0 or C0 \& C1)\\ scalar: $10^{-6}$\end{tabular} & $10^{-5}$ \\
      batch size & {32} \\
      Hessian batch size & {32} \\
      learning rate & {0.0005} \\
      initial $\lambda$ & {0.01} \\
      \bottomrule
  \end{tabular}
\end{table}

\begin{table}
  \centering
	\caption{\textbf{Computational requirements of each operation}. We provide computational, memory complexity of each step.}
	\label{tab:complexity}
	\begin{tabular}{lll}
		\toprule
		& \textbf{\begin{tabular}[c]{@{}l@{}}Computational\\complexity\end{tabular}} & \textbf{\begin{tabular}[c]{@{}l@{}}Memory\\ complexity\end{tabular}}\\ \midrule
		\textsc{predict} &&\\
    \hspace{0.5cm}$\circ$ inverse of sum (\cref{alg:low_rank_filtering} L3) &$\bigO (k^2D + k^3)$&$\bigO (kD  + k^2)$\\
    \hspace{0.5cm}$\circ$ update $\vm_t^-$ (\cref{alg:low_rank_filtering} L3) &$\bigO (k^2 + kD)$&$\bigO (k^2 + kD)$\\
    \textsc{update} &&\\
    \hspace{0.5cm}$\circ$ compute GGN (\cref{alg:low_rank_filtering} L6) &$\bigO (C^2 + CD)$&$\bigO (C^2+ CD)$\\
    \hspace{0.5cm}$\circ$ square-root structure (\cref{alg:low_rank_filtering} L6) &$\bigO ((k^2+C^2B)D)$&$\bigO (D(B+1)k + k^2)$\\
    \textsc{smooth} &&\\
    \hspace{0.5cm}$\circ$ compute $\mG_t$ (\cref{eq:lgf:smoothing:mean-and-cov}) &$\bigO (k^2D + k^3)$&$\bigO (kD + k^2)$\\
    \hspace{0.5cm}$\circ$ update $\vm_t^s$ (\cref{eq:lgf:smoothing:mean-and-cov})&$\bigO (k^2 + kD)$&$\bigO (k^2 + kD)$\\
		
    \bottomrule
	\end{tabular}
\end{table}

\subsection{Experimental Results}
\label{app:exp_details_results}

We provide additional experimental results below.

\begin{figure}[t]
	\centering
	\includegraphics[width=0.99\linewidth]{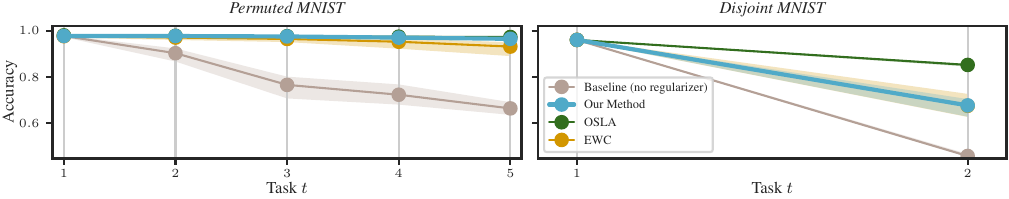}
	\caption{\textbf{Comparing our diagonal plus low-rank GGN approximation to other regularizers}. Mean performance of the current model on all previously encountered tasks (shaded areas are the min/max across $8$ seeds). For \emph{Permuted} and \emph{Disjoint \mnist} we observe that our diagonal plus low-rank GGN approximation (\colorline{TUlightblue}) leads to significantly lower rates of forgetting compared to no regularization (\colorline{TUmauve}). It also tends to be slightly better than EWC (\colorline{TUorange}) and comparable to OSLA (\colorline{TUdarkgreen}).}
	\label{fig:benchmark}
\end{figure}

\begin{figure}[t]
	\centering
	\includegraphics[width=0.99\linewidth]{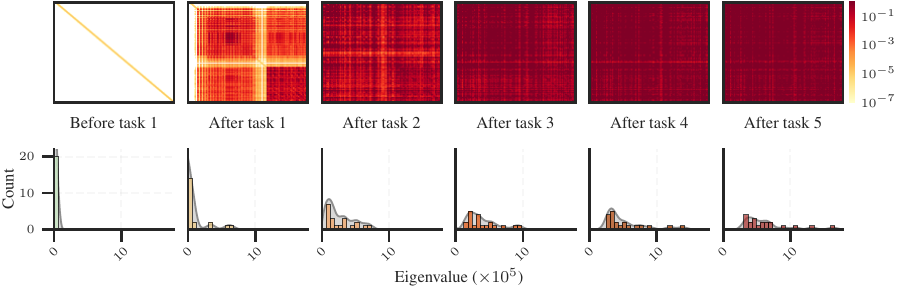}
	\caption{\textbf{The low-rank approximation across training (using rank $\mathbf{k\!=\!20}$).} \textit{(Top)} Our rank $k\!=\!20$ approximation to the precision matrix across tasks $t$ on \textit{Permuted} \mnist (see \cref{fig:eigs} for $k\!=\!10$). For better readability, values below $10^{-7}$ are shown in white. \textit{(Bottom)} Histograms of the eigenvalues of the approximation's low-rank part. With growing $t$, the eigenvalues increase in magnitude, indicating larger certainty and less flexibility.}
	\label{fig:eigs_20}
\end{figure}

  \begin{figure}
	\centering
	\includegraphics[width=0.99\textwidth]{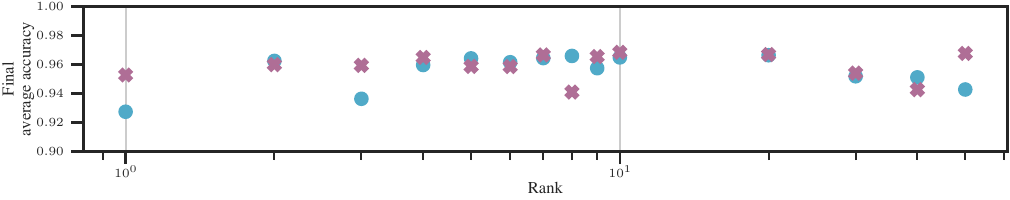}
	\caption{\textbf{Comparison of final average accuracy on \emph{Permuted} \mnist for different ranks $k$ of our diagonal plus low-rank approximation}. We show the final average accuracy for two different seeds (\colordot{TUlightblue} and \colorcross{TUpurple}) when using varying ranks for the low-rank part. We observe that already a very low rank $k\!\geq\!2$ provides a competitive performance. This indicates that a low rank is sufficient to provide a meaningful yet cheap curvature approximation. Increasing the rank beyond $\approx\!10$ (the number of classes) offers little additional benefit and can even slightly degrade performance.}
	\label{fig:ranks}
\end{figure}

\begin{figure}[t]
	\centering
	\includegraphics[width=0.99\linewidth]{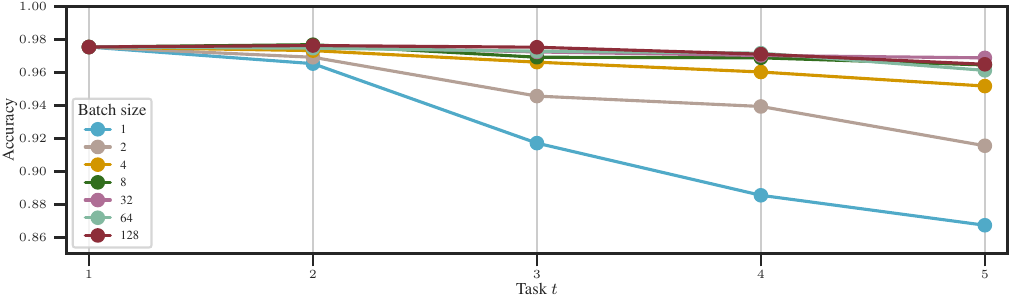}
	\caption{\textbf{The effect of the Hessian's batch size on the performance}. Using larger batch sizes to compute the GGN approximation of the Hessian help with the curvature estimation. Consequently, larger batch sizes also tend to improve the performance across tasks. However, we observe that a relatively modest and thus cheap batch size of $8$ is sufficient to provide competitive performance.}
  \label{fig:bs_hes}
\end{figure}

\begin{figure}
	\centering
	\includegraphics[width=0.99\textwidth]{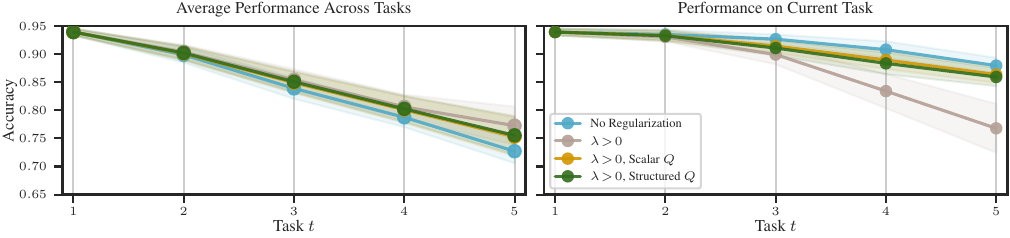}
	\caption{\textbf{The effect of $\mQ$ on the average and current task's performance (across seeds).} \textit{(Left)} Looking at the average performance across currently observed tasks, we see that without regularization (\colorline{TUlightblue}), performance significantly drops. With regularization (and possibly $\mQ\!>\!0$) (\eg \colorline{TUmauve}, \colorline{TUorange}, \colorline{TUdarkgreen}), we can boost the average performance to roughly similar levels.
	\textit{(Right)} Crucially, adding $\mQ\!>\!0$ allows us to boost the performance on the current task, while maintaining a similar average performance. Adding a structured $\mQ$ (here, one that specifically targets the first two convolutional layers) (see \Cref{sec:exp_Q}), tends too perform slightly stronger in terms of average performance across our experiments.
	Shaded areas show $\pm$ one standard deviation across $8$ random seeds.}
	\label{fig:Q_seeds}
\end{figure}

%% file: appendix/bibliography.bib
@String{ARXIV       = {arXiv}}

@String{ICLR        = {International Conference on Learning Representations (ICLR)}}

@String{ICML        = {International Conference on Machine Learning (ICML)}}

@String{IEEE        = {IEEE}}

@String{MIT         = {MIT Press}}

@String{Springer    = {Springer}}

@article{Kirkpatrick2017,
  author =       {Kirkpatrick, James and Pascanu, Razvan and Rabinowitz, Neil
                  and Veness, Joel and Desjardins, Guillaume and Rusu, Andrei A.
                  and Milan, Kieran and Quan, John and Ramalho, Tiago and
                  Grabska-Barwinska, Agnieszka and Hassabis, Demis and Clopath,
                  Claudia and Kumaran, Dharshan and Hadsell, Raia},
  title =        {Overcoming catastrophic forgetting in neural networks},
  journal =      {Proceedings of the National Academy of Sciences},
  year =         2017,
  volume =       114,
  number =       13,
  pages =        {3521--3526},
  publisher =    {Proceedings of the National Academy of Sciences}
}

@inproceedings{titsias2020functional,
  title =        {Functional Regularisation for Continual Learning with
                  {G}aussian Processes},
  author =       {Michalis K. Titsias and Jonathan Schwarz and Alexander G. de
                  G. Matthews and Razvan Pascanu and Yee Whye Teh},
  booktitle =    {International Conference on Learning Representations},
  year =         2020,
  url =          {https://openreview.net/forum?id=HkxCzeHFDB}
}

@inproceedings{nguyen2018variational,
  title =        {Variational Continual Learning},
  author =       {Cuong V. Nguyen and Yingzhen Li and Thang D. Bui and Richard
                  E. Turner},
  booktitle =    {International Conference on Learning Representations},
  year =         2018,
  url =          {https://openreview.net/forum?id=BkQqq0gRb},
}

@inproceedings{zenke2017continual,
  author =       {Zenke, Friedemann and Poole, Ben and Ganguli, Surya},
  title =        {Continual learning through synaptic intelligence},
  year =         2017,
  publisher =    {JMLR.org},
  booktitle =    {Proceedings of the 34th International Conference on Machine
                  Learning - Volume 70},
  pages =        {3987–3995},
  numpages =     9,
  location =     {Sydney, NSW, Australia},
  series =       {ICML'17}
}

@book{book,
  title =        {Lifelong Machine Learning},
  author =       {Zhiyuan Chen and Bing Liu},
  year =         2018,
  publisher =    {Springer}
}

@article{klasson2022learn,
  title =        {Learn the Time to Learn: Replay Scheduling in Continual
                  Learning},
  author =       {Marcus Klasson and Hedvig Kjellstrom and Cheng Zhang},
  journal =      {Transactions on Machine Learning Research},
  issn =         {2835-8856},
  year =         2023,
  url =          {https://openreview.net/forum?id=Q4aAITDgdP}
}

@inproceedings{rolnick2019experience,
  author =       {Rolnick, David and Ahuja, Arun and Schwarz, Jonathan and
                  Lillicrap, Timothy and Wayne, Gregory},
  booktitle =    {Advances in Neural Information Processing Systems},
  editor =       {H. Wallach and H. Larochelle and A. Beygelzimer and F.
                  d\textquotesingle Alch\'{e}-Buc and E. Fox and R. Garnett},
  publisher =    {Curran Associates, Inc.},
  title =        {Experience Replay for Continual Learning},
  url =
                  {https://proceedings.neurips.cc/paper_files/paper/2019/file/fa7cdfad1a5aaf8370ebeda47a1ff1c3-Paper.pdf},
  volume =       32,
  year =         2019
}

@book{datashift_book,
  author =       {Quionero-Candela, Joaquin and Sugiyama, Masashi and
                  Schwaighofer, Anton and Lawrence, Neil D.},
  title =        {Dataset Shift in Machine Learning},
  year =         2009,
  isbn =         0262170051,
  publisher =    {The MIT Press},
}

@ARTICLE{review_cl,
  author =       {Wang, Liyuan and Zhang, Xingxing and Su, Hang and Zhu, Jun},
  journal =      {IEEE Transactions on Pattern Analysis and Machine
                  Intelligence},
  title =        {A Comprehensive Survey of Continual Learning: Theory, Method
                  and Application},
  year =         2024,
  pages =        {1-20},
  doi =          {10.1109/TPAMI.2024.3367329}
}

@inproceedings{daxberger2022laplace,
  title =        {{L}aplace Redux - Effortless {B}ayesian Deep Learning},
  author =       {Erik Daxberger and Agustinus Kristiadi and Alexander Immer and
                  Runa Eschenhagen and Matthias Bauer and Philipp Hennig},
  booktitle =    {Advances in Neural Information Processing Systems},
  editor =       {A. Beygelzimer and Y. Dauphin and P. Liang and J. Wortman
                  Vaughan},
  year =         2021,
  url =          {https://openreview.net/forum?id=gDcaUj4Myhn}
}

@inproceedings{martens2020optimizing,
  author =       {Martens, James and Grosse, Roger},
  title =        {Optimizing neural networks with Kronecker-factored approximate
                  curvature},
  year =         2015,
  publisher =    {JMLR.org},
  booktitle =    {Proceedings of the 32nd International Conference on
                  International Conference on Machine Learning - Volume 37},
  pages =        {2408–2417},
  numpages =     10,
  location =     {Lille, France},
  series =       {ICML'15}
}

@InProceedings{chang2023lowrank,
  author    = {Chang, Peter G. and Dur\'an-Mart\'in, Gerardo and Shestopaloff, Alex and Jones, Matt and Murphy, Kevin Patrick},
  title     = {Low-rank extended {K}alman filtering for online learning of neural networks from streaming data},
  booktitle = {Proceedings of The 2nd Conference on Lifelong Learning Agents},
  year      = {2023},
  editor    = {Chandar, Sarath and Pascanu, Razvan and Sedghi, Hanie and Precup, Doina},
  volume    = {232},
  series    = {Proceedings of Machine Learning Research},
  publisher = {PMLR},
  month     = {22--25 Aug},
  pages     = {1025--1071},
  url       = {https://proceedings.mlr.press/v232/chang23a.html},
}

@article{verwimp2024continual,
  title =        {Continual Learning: Applications and the Road Forward},
  author =       {Eli Verwimp and Rahaf Aljundi and Shai Ben-David and Matthias
                  Bethge and Andrea Cossu and Alexander Gepperth and Tyler L.
                  Hayes and Eyke H{\"u}llermeier and Christopher Kanan and
                  Dhireesha Kudithipudi and Christoph H. Lampert and Martin
                  Mundt and Razvan Pascanu and Adrian Popescu and Andreas S.
                  Tolias and Joost van de Weijer and Bing Liu and Vincenzo
                  Lomonaco and Tinne Tuytelaars and Gido M van de Ven},
  journal =      {Transactions on Machine Learning Research},
  issn =         {2835-8856},
  year =         2024,
  url =          {https://openreview.net/forum?id=axBIMcGZn9}
}

@inproceedings{Ritter2018,
  author =       {Ritter, Hippolyt and Botev, Aleksandar and Barber, David},
  title =        {Online structured {L}aplace approximations for overcoming
                  catastrophic forgetting},
  year =         2018,
  publisher =    {Curran Associates Inc.},
  address =      {Red Hook, NY, USA},
  booktitle =    {Proceedings of the 32nd International Conference on Neural
                  Information Processing Systems},
  pages =        {3742–3752},
  numpages =     11,
  location =     {Montr\'{e}al, Canada},
  series =       {NIPS'18}
}

@misc{koh2021wilds,
  title =        {WILDS: A Benchmark of in-the-Wild Distribution Shifts},
  author =       {Pang Wei Koh and Shiori Sagawa and Henrik Marklund and Sang
                  Michael Xie and Marvin Zhang and Akshay Balsubramani and
                  Weihua Hu and Michihiro Yasunaga and Richard Lanas Phillips
                  and Irena Gao and Tony Lee and Etienne David and Ian Stavness
                  and Wei Guo and Berton A. Earnshaw and Imran S. Haque and Sara
                  Beery and Jure Leskovec and Anshul Kundaje and Emma Pierson
                  and Sergey Levine and Chelsea Finn and Percy Liang},
  year =         2021,
  eprint =       {2012.07421},
  archiveprefix ={arXiv},
  primaryclass = {cs.LG}
}

@incollection{McCloskey1989,
  author =       {McCloskey, Michael and Cohen, Neal J.},
  title =        {Catastrophic Interference in Connectionist Networks: The
                  Sequential Learning Problem},
  year =         1989,
  editor =       {Gordon H. Bower},
  volume =       24,
  booktitle =    {Psychology of Learning and Motivation},
  publisher =    {Academic Press},
  pages =        {109-165}
}

@article{MacKay1992,
  author =       {MacKay, David J. C.},
  title =        {A Practical {B}ayesian Framework for Backpropagation Networks},
  journal =      {Neural Computation},
  year =         1992,
  volume =       4,
  number =       3,
  month =        05,
  pages =        {448-472}
}

@book{Sarkka_Svensson_2023,
  place =        {Cambridge},
  edition =      2,
  series =       {Institute of Mathematical Statistics Textbooks},
  title =        {{B}ayesian Filtering and Smoothing},
  publisher =    {Cambridge University Press},
  author =       {Särkkä, Simo and Svensson, Lennart},
  year =         2023,
  collection =   {Institute of Mathematical Statistics Textbooks}
}

@article{koyama2010,
  title =        {Approximate methods for state-space models},
  author =       {Koyama, Shinsuke and Castellanos P{\'e}rez-Bolde, Lucia and
                  Shalizi, Cosma Rohilla and Kass, Robert E},
  journal =      {Journal of the American Statistical Association},
  volume =       105,
  number =       489,
  pages =        {170--180},
  year =         2010,
  publisher =    {Taylor \& Francis}
}

@article{kalman1960,
  author =       {Kalman, R. E.},
  title =        {A New Approach to Linear Filtering and Prediction Problems},
  journal =      {Journal of Basic Engineering},
  volume =       82,
  number =       1,
  pages =        {35-45},
  year =         1960,
  month =        03,
  issn =         {0021-9223},
  doi =          {10.1115/1.3662552},
  url =          {https://doi.org/10.1115/1.3662552},
}

@article{KalmanBucy1961,
  title =        {New Results in Linear Filtering and Prediction Theory},
  author =       {Rudolf E. Kalman and Richard S. Bucy},
  journal =      {Journal of Basic Engineering},
  year =         1961,
  volume =       83,
  pages =        {95-108},
  url =          {https://api.semanticscholar.org/CorpusID:8141345}
}

@article{rauchtungstriebel1965,
  author =       {Rauch, H. E. and Tung, F. and Striebel, C. T.},
  title =        {Maximum likelihood estimates of linear dynamic systems},
  journal =      {AIAA Journal},
  volume =       3,
  number =       8,
  pages =        {1445-1450},
  year =         1965,
  doi =          {10.2514/3.3166},
  URL =          {https://doi.org/10.2514/3.3166},
  eprint =       {https://doi.org/10.2514/3.3166}
}

@book{jazwinski2007stochastic,
  title =        {Stochastic processes and filtering theory},
  author =       {Jazwinski, Andrew H},
  year =         2007,
  publisher =    {Courier Corporation}
}

@ARTICLE{Uhlmann2000,
  author =       {Julier, S. and Uhlmann, J. and Durrant-Whyte, H.F.},
  journal =      {IEEE Transactions on Automatic Control},
  title =        {A new method for the nonlinear transformation of means and
                  covariances in filters and estimators},
  year =         2000,
  volume =       45,
  number =       3,
  pages =        {477-482},
  doi =          {10.1109/9.847726}
}

@inproceedings{titsias2024kalman,
  title =        {{K}alman Filter for Online Classification of Non-Stationary
                  Data},
  author =       {Michalis Titsias and Alexandre Galashov and Amal Rannen-Triki
                  and Razvan Pascanu and Yee Whye Teh and Jorg Bornschein},
  booktitle =    {The Twelfth International Conference on Learning
                  Representations},
  year =         2024,
  url =          {https://openreview.net/forum?id=ZzmKEpze8e}
}

@inproceedings{chang2022on,
  title =        {On diagonal approximations to the extended {K}alman filter for
                  online training of {B}ayesian neural networks},
  author =       {Peter G. Chang and Kevin Patrick Murphy and Matt Jones},
  booktitle =    {Continual Lifelong Learning Workshop at ACML 2022},
  year =         2022,
  url =          {https://openreview.net/forum?id=asgeEt25kk}
}

@INPROCEEDINGS{puskorius1991,
  author =       {Puskorius, G.V. and Feldkamp, L.A.},
  booktitle =    {IJCNN-91-Seattle International Joint Conference on Neural
                  Networks},
  title =        {Decoupled extended {K}alman filter training of feedforward
                  layered networks},
  year =         1991,
  volume =       {i},
  pages =        {771-777 vol.1},
  doi =          {10.1109/IJCNN.1991.155276}
}

@inproceedings{singhal1988,
  author =       {Singhal, Sharad and Wu, Lance},
  booktitle =    {Advances in Neural Information Processing Systems},
  editor =       {D. Touretzky},
  publisher =    {Morgan-Kaufmann},
  title =        {Training Multilayer Perceptrons with the Extended {K}alman
                  Algorithm},
  url =
                  {https://proceedings.neurips.cc/paper_files/paper/1988/file/38b3eff8baf56627478ec76a704e9b52-Paper.pdf},
  volume =       1,
  year =         1988
}

@article{Feldkamp1998,
  title =        {Enhanced multi-stream Kalman filter training for recurrent
                  networks},
  author =       {Feldkamp, Lee A and Prokhorov, Danil V and Eagen, Charles F
                  and Yuan, Fumin},
  journal =      {Nonlinear modeling: advanced black-box techniques},
  pages =        {29--53},
  year =         1998,
  publisher =    {Springer}
}

@ARTICLE{Bell1993,
  author =       {B. M. Bell and F. W. Cathey},
  title =        {The iterated {K}alman filter update as a {G}auss--{N}ewton
                  method},
  journal =      {IEEE Transaction on Automatic Control},
  year =         1993,
  volume =       38,
  number =       2,
  pages =        {294--297}
}

@ARTICLE{Schraudolph2002,
  author =       {Schraudolph, Nicol N.},
  journal =      {Neural Computation},
  title =        {Fast Curvature Matrix-Vector Products for Second-Order
                  Gradient Descent},
  year =         2002,
  volume =       14,
  number =       7,
  pages =        {1723-1738},
  doi =          {10.1162/08997660260028683}
}

@INPROCEEDINGS{8545895,
  author={Liu, Xialei and Masana, Marc and Herranz, Luis and Van de Weijer, Joost and López, Antonio M. and Bagdanov, Andrew D.},
  booktitle={2018 24th International Conference on Pattern Recognition (ICPR)}, 
  title={Rotate your Networks: Better Weight Consolidation and Less Catastrophic Forgetting}, 
  year={2018},
  volume={},
  number={},
  pages={2262-2268},
  doi={10.1109/ICPR.2018.8545895}}

@InProceedings{pmlr-v80-schwarz18a,
  title = 	 {Progress \& Compress: A scalable framework for continual learning},
  author =       {Schwarz, Jonathan and Czarnecki, Wojciech and Luketina, Jelena and Grabska-Barwinska, Agnieszka and Teh, Yee Whye and Pascanu, Razvan and Hadsell, Raia},
  booktitle = 	 {Proceedings of the 35th International Conference on Machine Learning},
  pages = 	 {4528--4537},
  year = 	 {2018},
  editor = 	 {Dy, Jennifer and Krause, Andreas},
  volume = 	 {80},
  series = 	 {Proceedings of Machine Learning Research},
  month = 	 {10--15 Jul},
  publisher =    {PMLR},
}

@INPROCEEDINGS {9009502,
author = {D. Park and S. Hong and B. Han and K. Lee},
booktitle = {2019 IEEE/CVF International Conference on Computer Vision (ICCV)},
title = {Continual Learning by Asymmetric Loss Approximation With Single-Side Overestimation},
year = {2019},
volume = {},
issn = {},
pages = {3334-3343},
doi = {10.1109/ICCV.2019.00343},
url = {https://doi.ieeecomputersociety.org/10.1109/ICCV.2019.00343},
publisher = {IEEE Computer Society},
address = {Los Alamitos, CA, USA},
month = {nov}
}

@inproceedings{NIPS2017_f708f064,
 author = {Lee, Sang-Woo and Kim, Jin-Hwa and Jun, Jaehyun and Ha, Jung-Woo and Zhang, Byoung-Tak},
 booktitle = {Advances in Neural Information Processing Systems},
 editor = {I. Guyon and U. Von Luxburg and S. Bengio and H. Wallach and R. Fergus and S. Vishwanathan and R. Garnett},
 pages = {},
 publisher = {Curran Associates, Inc.},
 title = {Overcoming Catastrophic Forgetting by Incremental Moment Matching},
 url = {https://proceedings.neurips.cc/paper_files/paper/2017/file/f708f064faaf32a43e4d3c784e6af9ea-Paper.pdf},
 volume = {30},
 year = {2017}
}

@Misc{Dohare2023,
  author        = {Dohare, Shibhansh and Hernandez-Garcia, J. Fernando and Rahman, Parash and Mahmood, A. Rupam and Sutton, Richard S.},
  title         = {{Maintaining Plasticity in Deep Continual Learning}},
  year          = {2023},
  eprint        = {2306.13812},
  archiveprefix = {arXiv},
}

@article{deng2012mnist,
  title={The mnist database of handwritten digit images for machine learning research},
  author={Deng, Li},
  journal={IEEE Signal Processing Magazine},
  volume={29},
  number={6},
  pages={141--142},
  year={2012},
  publisher={IEEE}
}

@software{jax2018github,
  author = {James Bradbury and Roy Frostig and Peter Hawkins and Matthew James Johnson and Chris Leary and Dougal Maclaurin and George Necula and Adam Paszke and Jake Vander{P}las and Skye Wanderman-{M}ilne and Qiao Zhang},
  title = {{JAX}: composable transformations of {P}ython+{N}um{P}y programs},
  url = {http://github.com/google/jax},
  version = {0.3.13},
  year = {2018},
}

@misc{munoz2019,
      title={Continual Multi-task Gaussian Processes}, 
      author={Pablo Moreno-Muñoz and Antonio Artés-Rodríguez and Mauricio A. Álvarez},
      year={2019},
      eprint={1911.00002},
      archivePrefix={arXiv}
}

@InProceedings{kapoor2021,
  title = 	 {Variational Auto-Regressive Gaussian Processes for Continual Learning},
  author =       {Kapoor, Sanyam and Karaletsos, Theofanis and Bui, Thang D},
  booktitle = 	 {Proceedings of the 38th International Conference on Machine Learning},
  pages = 	 {5290--5300},
  year = 	 {2021},
  editor = 	 {Meila, Marina and Zhang, Tong},
  volume = 	 {139},
  series = 	 {Proceedings of Machine Learning Research},
  month = 	 {18--24 Jul},
  publisher =    {PMLR},
  url = 	 {https://proceedings.mlr.press/v139/kapoor21b.html}
}

@inproceedings{Titsias2020,
  author       = {Michalis K. Titsias and
                  Jonathan Schwarz and
                  Alexander G. de G. Matthews and
                  Razvan Pascanu and
                  Yee Whye Teh},
  title        = {Functional Regularisation for Continual Learning with Gaussian Processes},
  booktitle    = {8th International Conference on Learning Representations, {ICLR} 2020,
                  Addis Ababa, Ethiopia, April 26-30, 2020},
  publisher    = {OpenReview.net},
  year         = {2020},
  url          = {https://openreview.net/forum?id=HkxCzeHFDB},
  bibsource    = {dblp computer science bibliography, https://dblp.org}
}

@misc{mirzadeh2020,
      title={Linear Mode Connectivity in Multitask and Continual Learning}, 
      author={Seyed Iman Mirzadeh and Mehrdad Farajtabar and Dilan Gorur and Razvan Pascanu and Hassan Ghasemzadeh},
      year={2021},
      booktitle = {International Conference on Learning
                  Representations},
      eprint={2010.04495},
      archivePrefix={arXiv},
      primaryClass={cs.LG},
      url={https://arxiv.org/abs/2010.04495}, 
}

@article{adel2024,
author = {Adel, Tameem},
year = {2024},
month = {06},
pages = {377-417},
title = {Similarity-Based Adaptation for Task-Aware and Task-Free Continual Learning},
volume = {80},
journal = {Journal of Artificial Intelligence Research},
doi = {10.1613/jair.1.15693}
}

@misc{ramesh2022,
      title={Model Zoo: A Growing "Brain" That Learns Continually}, 
      author={Rahul Ramesh and Pratik Chaudhari},
      year={2022},
      eprint={2106.03027},
      archivePrefix={arXiv},
      booktitle = {International Conference on Learning
                  Representations},
      primaryClass={cs.LG},
      url={https://arxiv.org/abs/2106.03027}, 
}

@inproceedings{wang2022,
      author = {Wang, Liyuan and Zhang, Xingxing and Li, Qian and Zhu, Jun and Zhong, Yi},
      title = {CoSCL: Cooperation of Small Continual Learners is Stronger Than Big One},
      year = {2022},
      isbn = {978-3-031-19808-3},
      publisher = {Springer-Verlag},
      doi = {10.1007/978-3-031-19809-0_15},
      booktitle = {Computer Vision – ECCV 2022: 17th European Conference, Tel Aviv, Israel, October 23–27, 2022, Proceedings, Part XXVI},
      pages = {254–271},
      numpages = {18}
  }

@misc{EWC-mention,
      title={Three scenarios for continual learning}, 
      author={Gido M. van de Ven and Andreas S. Tolias},
      year={2019},
      eprint={1904.07734},
      archivePrefix={arXiv},
      primaryClass={cs.LG},
      url={https://arxiv.org/abs/1904.07734}, 
}

@misc{EWC-indepth,
      title={Elastic Weight Consolidation (EWC): Nuts and Bolts}, 
      author={Abhishek Aich},
      year={2021},
      eprint={2105.04093},
      archivePrefix={arXiv},
      primaryClass={cs.CV},
      url={https://arxiv.org/abs/2105.04093}, 
}

@misc{kristiadi2024soberlookllmsmaterial,
      title={A Sober Look at LLMs for Material Discovery: Are They Actually Good for Bayesian Optimization Over Molecules?}, 
      author={Agustinus Kristiadi and Felix Strieth-Kalthoff and Marta Skreta and Pascal Poupart and Alán Aspuru-Guzik and Geoff Pleiss},
      year={2024},
      eprint={2402.05015},
      archivePrefix={arXiv},
      primaryClass={cs.LG},
      url={https://arxiv.org/abs/2402.05015}, 
}
